\documentclass[sigconf, nonacm]{acmart}
\usepackage{enumitem}
\usepackage{xcolor} 
\usepackage{makecell}
\usepackage{mathtools} 
\usepackage{tikz}
\usepackage{pgfplots}
\usepackage{pgfplotstable}
\usepackage{subcaption}
\usepackage{hyperref}
\usepackage{flushend}
\usepackage{hyperref}
\usepackage{tikz}
\usetikzlibrary{calc}
\definecolor{lavenderlike}{HTML}{CCCCFF}

\hypersetup{
    colorlinks=true,
    linkcolor=blue,
    urlcolor=blue,
    citecolor=blue
}

\pgfplotsset{compat=1.16}

\usepgfplotslibrary{groupplots}
\pgfplotsset{compat=1.18}
\usetikzlibrary{patterns,shapes.geometric,calc,positioning,arrows.meta}

\newtheoremstyle{examplestyle} 
    {1pt} 
    {1pt} 
    {\itshape} 
    {} 
    {\bfseries} 
    {.} 
    {.1em} 
    {} 
\theoremstyle{examplestyle}
\newtheorem{example}{Example}

\newtheoremstyle{definitionstyle} 
    {1pt} 
    {1pt} 
    {\itshape} 
    {} 
    {\bfseries} 
    {.} 
    {.3em} 
    {} 
\theoremstyle{definitionstyle}
\newtheorem{definition}{Definition}

\newtheoremstyle{lemmastyle} 
    {1pt} 
    {1pt} 
    {\itshape} 
    {} 
    {\bfseries} 
    {.} 
    {.3em} 
    {} 
\theoremstyle{lemmastyle}
\newtheorem{lemma}{Lemma}

\newcommand{\warning}[1]{\textcolor{black}{#1}}
\newcommand{\warninge}[1]{\textcolor{black}{#1}}
\usepackage{multirow}  
\usepackage{array}     
\newcommand\vldbdoi{XX.XX/XXX.XX}
\newcommand\vldbpages{XXX-XXX}
\newcommand\vldbvolume{14}
\newcommand\vldbissue{1}
\newcommand\vldbyear{2024}
\newcommand\vldbauthors{\authors}
\newcommand\vldbtitle{\shorttitle} 
\newcommand\vldbavailabilityurl{URL_TO_YOUR_ARTIFACTS}
\newcommand\vldbpagestyle{plain} 

\begin{document}
\title{\textsf{MH-GIN}: Multi-scale Heterogeneous Graph-based Imputation Network for AIS Data (Extended Version)}

\author{Hengyu Liu$^1$ \quad Tianyi Li$^{1,\dagger}$ \quad Yuqiang He$^2$ \quad Kristian Torp$^1$ \quad Yushuai Li$^1$  \quad Christian S. Jensen$^1$}
\affiliation{
\institution{$^1$Department of Computer Science, Aalborg University, Denmark\\
$^2$School of Computer, Guangxi University, Guangxi, China}
$^1$\{heli, tianyi, torp, yusli, csj\}@cs.aau.dk, $^2$heyq1314@gmail.com}

\begin{abstract}
Location-tracking data from the Automatic Identification System, much of which is publicly available, plays a key role in a range of maritime safety and monitoring applications.  However, the data suffers from missing values that hamper downstream applications. Imputing the missing values is challenging because the values of different heterogeneous attributes are updated at diverse rates, resulting in the occurrence of multi-scale dependencies among attributes. Existing imputation methods that assume similar update rates across attributes are unable to capture and exploit such dependencies, limiting their imputation accuracy. We propose \textsf{MH-GIN}, a Multi-scale Heterogeneous Graph-based Imputation Network that aims improve imputation accuracy by capturing multi-scale dependencies. Specifically, \textsf{MH-GIN} first extracts multi-scale temporal features for each attribute while preserving their intrinsic heterogeneous characteristics. Then, it constructs a multi-scale heterogeneous graph to explicitly model dependencies between heterogeneous attributes to enable more accurate imputation of missing values through graph propagation. Experimental results on two real-world datasets find that \textsf{MH-GIN} is capable of an average 57\% reduction in imputation errors compared to state-of-the-art methods, while maintaining computational efficiency. The source code and implementation details of \textsf{MH-GIN} are publicly available\footnote{\href{https://github.com/hyLiu1994/MH-GIN}{\textcolor{blue}{https://github.com/hyLiu1994/MH-GIN}}}.
\end{abstract}
\maketitle

\pagestyle{\vldbpagestyle}
\begingroup\small\noindent\raggedright\textbf{PVLDB Reference Format:}\\
\vldbauthors. \vldbtitle. PVLDB, \vldbvolume(\vldbissue): \vldbpages, \vldbyear.\\
\href{https://doi.org/\vldbdoi}{doi:\vldbdoi}
\endgroup
\begingroup
\renewcommand\thefootnote{}\footnote{\noindent
$^{\dagger}$ Tianyi Li is the corresponding author.\\
This work is licensed under the Creative Commons BY-NC-ND 4.0 International License. Visit \url{https://creativecommons.org/licenses/by-nc-nd/4.0/} to view a copy of this license. For any use beyond those covered by this license, obtain permission by emailing \href{mailto:info@vldb.org}{info@vldb.org}. Copyright is held by the owner/author(s). Publication rights licensed to the VLDB Endowment. \\
\raggedright Proceedings of the VLDB Endowment, Vol. \vldbvolume, No. \vldbissue\ %
ISSN 2150-8097. \\
\href{https://doi.org/\vldbdoi}{doi:\vldbdoi} \\
}\addtocounter{footnote}{-1}\endgroup

\ifdefempty{\vldbavailabilityurl}{}{
\vspace{.3cm}
\begingroup\small\noindent\raggedright\textbf{PVLDB Artifact Availability:}\\
The source code, data, and/or other artifacts have been made available at \url{\vldbavailabilityurl}.
\endgroup
}

\section{Introduction}
The Automatic Identification System (AIS) is an automated tracking system that enhances navigational safety by enabling vessels to share their position, identification, and other essential information with nearby ships and coastal authorities~\cite{noauthor_methodological_2023}. AIS data is crucial for maritime stakeholders to operate applications spanning different domains, including vessel tracking, safety monitoring, logistics optimization, and trade forecasting. However, AIS data often suffer from quality issues, among which missing values represent a significant challenge. Missing values may be caused by a variety of various factors, such as equipment malfunctions, vessels switching off their transponders, or vessels being out of range of terrestrial receivers~\cite{Mekkaoui_automatic_2022}. Notably, statistics of data from the Danish Maritime Authority\footnote{\url{http://web.ais.dk/aisdata/}} reveal that missing rates in attributes range from 8\% to 83\%, which underscores the substantial need for data imputation in real-world scenarios. 
When AIS fails, downstream applications (such as vessel tracking~\cite{xie2025ais}, safety monitoring~\cite{vizion_ais_2025}, and navigation planning~\cite{spire_notall_2024}) often rely on traditional tools (such as radar and VHF radio~\cite{usermanual_USCoast_2024,usermanual_USCoast_2017}) or alternative systems (such as Long-Range Identification and Tracking (LRIT) and Vessel Monitoring Systems (VMS)~\cite{IMO_Long_range_2022}) typically at the cost of accuracy and efficiency.

\begin{figure}[!tbp]
\centering
\includegraphics[width=0.5\textwidth]{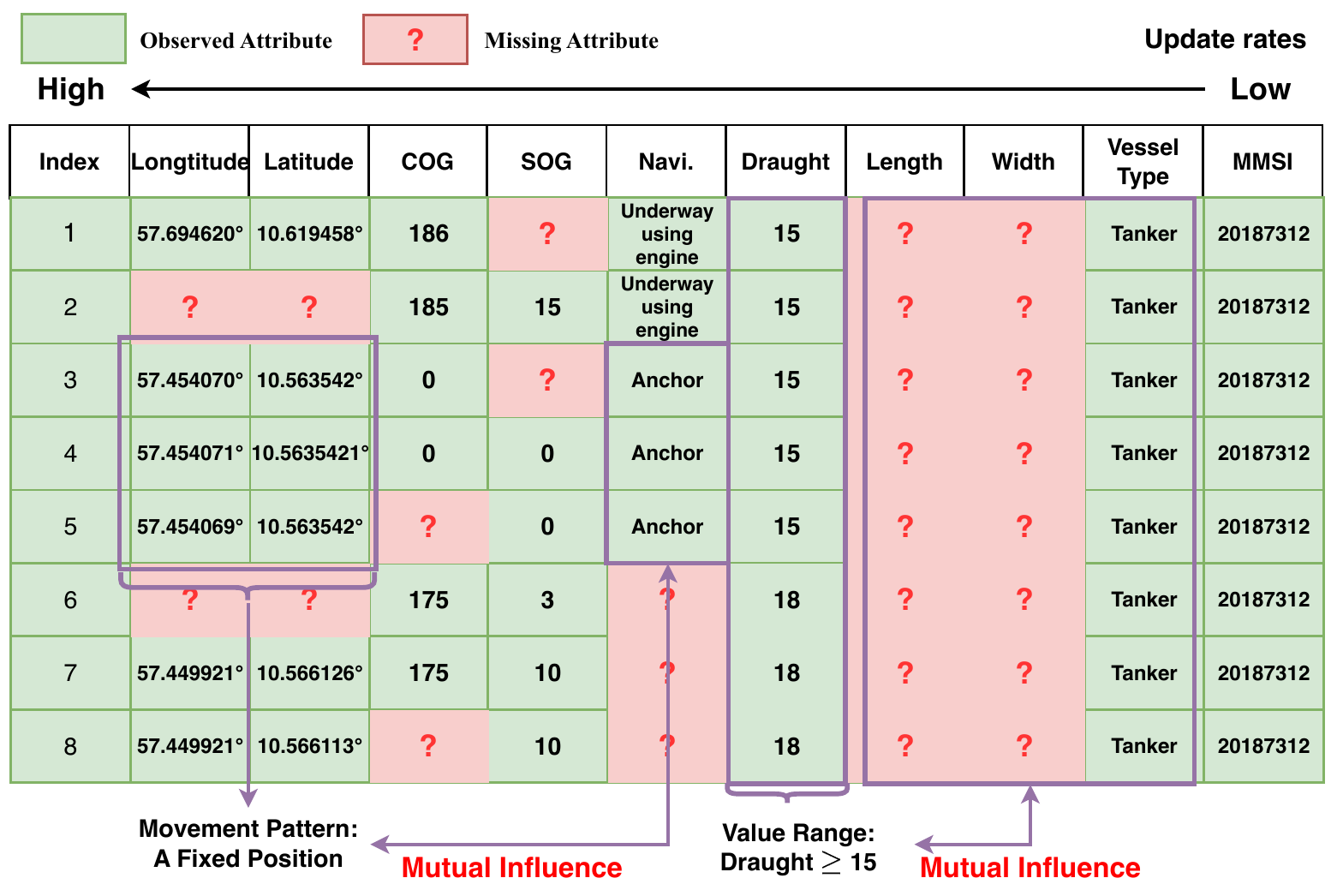}
\vspace{-4mm}
\caption{\warning{An example of AIS records from a vessel.}}
\vspace{-4mm}
\label{fig:toy_example}
\end{figure}

\begin{example} \label{ex:example_1}
Consider a vessel's AIS records during a port visit as illustrated in Figure~\ref{fig:toy_example}. The vessel follows a typical maritime pattern: entering the port (records 1--2), berthing at the terminal (records 3--5), and departing from the port (records 6--8). During this process, the vessel's Longitude, Latitude, COG, and SOG are updated at high frequency and exhibit random missing patterns. The vessel's Navigation Status and Draught are updated at low frequency and show block missing patterns. The vessel's Length, Width, and Vessel Type remain constant throughout, with Length and Width being entirely missing.
\end{example}

\warning{As shown in Example~\ref{ex:example_1}, AIS data exhibit three characteristics~\cite{sturgis_vessel_2022}: 1) There are \textbf{heterogeneous attributes} in AIS data, i.e., spatio-temporal attributes (Longtitude, Latitude), discrete attributes (Navigation Status), and continuous attributes (Draught).
2) Different attributes exhibit \textbf{diverse update rates}, leading to diverse missing patterns. As the update rate decreases, the missing pattern transitions from random to block to entirely missing.
3) There are \textbf{multi-scale dependencies} between attributes. Attributes at high time scales constrain the high-level features of attributes at low time scales without affecting their low-level features. Conversely, only the high-level features of attributes at low time scales influence attributes at higher time scales. For instance, as shown in Figure~\ref{fig:toy_example}, Navigation Status (Anchored) constrains a vessel's movement to a limited region around a fixed point (high-level feature: movement pattern of Longitude and Latitude) without specifying precise coordinates. Similarly, Vessel Type (Tanker) constrains the value range of Draught (high-level feature of Draught) without specifying the exact value. Conversely, the value range of Draught also constrains the value range of the vessel's Length and Width.}

\begin{table}[t]
    \centering
    \caption{\warninge{Summary of existing imputation methods}}
    \label{tab:related_work_comparison}
    \setlength{\tabcolsep}{2pt}\small
    \footnotesize
    \begin{tabular}{|l|c|c|c|}
    \hline
    \makecell[c]{Methodology \\ Class} & \makecell[c]{Heterogeneous \\ Attributes} & \makecell[c]{Diverse Update \\ Rates} & \makecell[c]{Multi-scale \\ Dependencies} \\
    \hline
    \makecell[c]{Heuristic \& \\ Statistical \\ Imputation} & Yes & No & No \\
    \hline
    \makecell[c]{Multi-variable \\ Time Series \\ Imputation} & \makecell[c]{No (Assumes \\ Homogeneity)} & \makecell[c]{No (Assumes \\ Uniform Update \\ Rates)} & \makecell[c]{No (Single-scale \\ Dependency)}  \\
    \hline
    \makecell[c]{Spatio-temporal \\ Imputation} & \makecell[c]{Partial (Focus on\\ Spatial and \\ Continuous \\ Attributes)} & \makecell[c]{Partial (Assumes \\ Uniform Rates \\ for Non-spatial)} & \makecell[c]{Partial (Single-scale \\ Dependency \\ for Non-spatial)}  \\
    \hline
    \makecell[c]{Trajectory \\ Imputation} & \makecell[c]{No (Spatio-temporal \\ Attributes Only)} & \makecell[c]{No (Assumes Similar \\ Update Rates)} & \makecell[c]{No (Single-scale \\ Dependency)} \\
    \hline
    \end{tabular}
\end{table}

However, existing imputation methods cannot effectively handle these characteristics. As shown in Table~\ref{tab:related_work_comparison}, existing imputation methods can be categorized into four categories, each with distinct limitations: 
1) Traditional and statistical methods~\cite{wothke2000longitudinal,mcknight2007missing,batista2002study,song_enriching_2020,acuna2004treatment,amiri2016missing,khayati_mind_2020} fall short due to their assumptions of linear relationships and their inability to handle diverse update rates and model multi-scale dependencies. 
2) Multi-variable time series imputation methods~\cite{che_recurrent_2018,du_saits_2023,tashiro2021csdi,cini_filling_2022} assume uniform update rates and treat attributes homogeneously. 
3) Spatio-temporal imputation methods~\cite{liu_pristi_2023,nie2024imputeformer,marisca_learning_nodate} assume static spatial relationships and uniform update rates for non-spatial attributes. 
4) Trajectory imputation methods~\cite{DBLP:journals/tmc/SiYXWLZTC24,nguyen2018multi,zhang2024long} focus exclusively on reconstructing vessel positions and rarely consider other heterogeneous attributes critical for comprehensive AIS analysis.
Effectively handling the characteristics of AIS data is challenging and presents significant obstacles.


\noindent \textit{\textbf{C1: How to effectively represent and simultaneously impute heterogeneous attributes?}} AIS data consist of four attribute types: spatio-temporal, cyclical, continuous, and discrete, and each has distinct intrinsic characteristics. The heterogeneous attributes require unified yet type-specific representation and imputation strategies, capable of preserving their individual properties. Such strategies are an essential precondition for capturing multi-scale dependencies and accurately imputing missing data. Existing imputation methods~\cite{cini_filling_2022, tashiro2021csdi, marisca_learning_nodate, nie2024imputeformer, zhang2024long} exhibit significant limitations in addressing this heterogeneity. They either focus exclusively on the spatio-temporal data or treat all attributes homogeneously, neglecting the unique properties of different attribute types.

\noindent \textit{\textbf{C2: How to effectively extract multi-scale temporal features from each attribute?}} As discussed earlier, AIS attributes operate at different time scales and influence each other indirectly. To capture the direct interactions among attributes at a particular scale accurately, it is essential to model temporal features individually at their own and higher scales. However, existing imputation methods~\cite{tashiro2021csdi, cini_filling_2022, cao2018brits, du_saits_2023} typically assume similar update rates across all attributes, thus extracting temporal features indiscriminately at all scales into a single, uninterpretable feature vector.

\noindent \textit{\textbf{C3: How to effectively model multi-scale dependencies between attributes?}} Multi-scale dependencies are complex, involving both dependencies between attributes at the same time scale and dependencies between attributes across different time scales. Effectively modeling and leveraging these dependencies for accurate imputation is challenging. Current methods~\cite{che_recurrent_2018, yoon2018estimating, du_saits_2023, bansal_missing_2023} do not differentiate direct and indirect attribute interactions at varying scales due to their similar update rate assumption, resulting in oversimplified attribute relationships and reduced imputation performance.

To address these challenges, we propose \textsf{MH-GIN}, a Multi-scale Heterogeneous Graph-based Imputation Network for AIS data.

\noindent \textit{\textbf{Addressing C1:}} \textsf{MH-GIN} utilize type-specific encoding and decoding strategies for four distinct attribute types: spatio-temporal, cyclical, continous, and discrete attributes, while preserving their intrinsic characteristics, such as spatial constraints, smooth wrap-around at boundaries.

\noindent \textit{\textbf{Addressing C2:}} \textsf{MH-GIN} adopts a hierarchical temporal feature extractor based on Deep Echo State Networks, enabling the capture of hierarchical temporal features for each attribute through progressive abstraction and leaky integration.

\noindent \textit{\textbf{Addressing C3:}} \textsf{MH-GIN} constructs a multi-scale heterogeneous graph to explicitly model two types of multi-scale dependencies. Specifically, the graph contains two kinds of subgraphs: time-scale subgraphs and attribute subgraphs. The former model dependencies between features of different attributes at the same scale, whereas the latter model dependencies across different scales within each attribute. Subsequently, \textsf{MH-GIN} imputes missing features through a two-stage graph propagation process, first aligning features within time-scale subgraphs, and then impute missing features through propagation within attribute subgraphs. 

Our main contributions are summarized as follows:
\setlength{\parsep}{0pt}
\begin{itemize}[itemsep=2pt, leftmargin=10pt]
\item We propose \textsf{MH-GIN}, a Multi-scale Heterogeneous Graph-based Imputation Network for AIS data, which captures multi-scale dependencies among heterogeneous attributes to enhance imputation performance.

\item We design complementary modules, including type-specific encoders and decoders and a hierarchical temporal feature extractor, to collaboratively extract hierarchical temporal features for each attribute type while preserving intrinsic characteristics of each attribute type.

\item We introduce a multi-scale dependency mining module, which constructs a multi-scale heterogeneous graph to model dependencies between attributes explicitly and performs imputation via two-stage graph propagation.

\item We evaluate \textsf{MH-GIN} extensively on two real-world AIS datasets, finding that \textsf{MH-GIN} is able to outperform state-of-the-art approaches, achieving an average 57\% reduction in imputation error while maintaining computational efficiency.
\end{itemize}

The remainder of the paper is structured as follows: 
Section~\ref{sec:related_work} reviews related work. 
Section~\ref{sec:preliminaries} presents preliminaries and formulates the problem. 
Section~\ref{sec:methodology} details the proposed \textsf{MH-GIN}. 
Section~\ref{sec:experiment} reports experimental results. 
Section~\ref{sec:conclusion} concludes the paper and outlines directions for future work.
\section{Related Work} 
\label{sec:related_work} 
Existing imputation methods can be categorized into four categories, as shown in Table~\ref{tab:related_work_comparison}, each with distinct limitations: 

\noindent \textbf{Traditional and Statistical Imputation Methods.} Traditional approaches for handling missing data include deletion methods, neighbor-based methods, constraint-based methods, and statistical imputation. Deletion methods, such as listwise deletion~\cite{wothke2000longitudinal} and pairwise deletion~\cite{mcknight2007missing}, directly remove data instances containing missing values, often resulting in significant information loss. Neighbor-based methods like KNN~\cite{batista2002study} impute missing values using local patterns derived from similar data points. Constraint-based methods~\cite{song_enriching_2020} leverage domain-specific rules, performing well when data adheres to particular constraints or patterns. Common statistical approaches include mean/median imputation~\cite{acuna2004treatment}, Last Observation Carried Forward (LOCF)\cite{amiri2016missing}, and linear interpolation\cite{khayati_mind_2020}. Despite being computationally efficient and straightforward, these traditional methods are inadequate for AIS data due to their implicit assumptions of linear relationships. AIS data inherently exhibits complex, non-linear patterns. Additionally, deletion-based approaches are particularly problematic for AIS, given its often extended periods of missingness. Thus, traditional methods struggle to represent the inherent multi-scale dependencies and heterogeneous attributes within AIS datasets.


\noindent \textbf{Multi-variable Time Series Imputation.}
Deep learning methods for multi-variable time series imputation have attracted considerable attention in recent years. Early RNN-based methods, such as GRU-D~\cite{che_recurrent_2018}, leverage gating mechanisms and decay factors to handle irregularly sampled data. More recently, transformer-based architectures, including SAITS~\cite{du_saits_2023} and DeepMVI~\cite{bansal_missing_2023}, utilize self-attention mechanisms to effectively capture long-range dependencies. Diffusion-based methods, such as CSDI~\cite{tashiro2021csdi}, employ conditional score-based diffusion models to produce high-quality imputations through iterative data denoising. Additionally, graph-based methods like GRIN~\cite{cini_filling_2022} represent each variable as a graph node, enabling message passing to capture cross-dimensional correlations. However, existing general-purpose imputation methods typically assume uniform update rates across all variables and treat attributes homogeneously. These assumptions conflict with the inherent characteristics of AIS data, which contains heterogeneous attributes operating at multiple time scales.


\noindent \textbf{Spatio-temporal Imputation.} Spatio-temporal imputation extends conventional time series methods by explicitly modeling spatial dependencies (e.g., sensor grids or geographic adjacency). Recent approaches incorporate graph neural networks (GNNs) or attention mechanisms to jointly learn spatio-temporal patterns. Representative methods include PriSTI~\cite{liu_pristi_2023}, which enhance diffusion-based models by capturing spatio-temporal dependencies and geographic relationships. ImputeFormer~\cite{nie2024imputeformer} utilizes a low-rank Transformer architecture to achieve a balance between inductive bias and model expressivity. SPIN~\cite{marisca_learning_nodate} proposes a spatio-temporal propagation framework with sparse attention, specifically designed to handle highly sparse observations by conditioning reconstruction exclusively on available data, thus preventing error propagation common in autoregressive GNNs. Although effective in sensor networks with relatively fixed spatial structures, these methods generally assume stationary spatial relationships or static graph topologies. However, AIS data involves vessels continuously moving across large maritime regions, leading to dynamic and evolving spatial relationships rather than static sensor locations. Therefore, existing spatio-temporal imputation methods face challenges in effectively modeling these dynamic spatial relationships. Furthermore, similar to multi-variable time series imputation methods, they typically assume uniform update rates for non-spatial attributes, rendering them ineffective in capturing diverse update frequencies inherent to AIS attributes.

%

\noindent \textbf{Trajectory Imputation.} Trajectory imputation methods reconstruct missing segments in the trajectories of moving objects (e.g., vehicles, pedestrians, and ships), typically leveraging motion continuity and spatio-temporal context. 
Transformer-based approaches, such as TrajBERT~\cite{DBLP:journals/tmc/SiYXWLZTC24}, embed geographic coordinates as tokens and reconstruct missing positions using bidirectional attention mechanisms. 
KAMEL~\cite{musleh2023kamel} maps trajectory imputation to the missing word problem in natural language processing, adapting BERT with spatial-awareness and multi-point imputation capabilities for scalable trajectory reconstruction. 
Recurrent and graph-based sequence models~\cite{xu2023uncovering, zhao2022trajgat, yoon2018estimating, lipton2016directly} exploit temporal dependencies and spatial constraints, capturing complex motion patterns and environmental context to impute missing trajectory segments. Meanwhile, generative models like GANs~\cite{shi2023tigan} and VAEs~\cite{qi2020imitative} produce plausible trajectories consistent with known movement behaviors. Specifically tailored maritime approaches include Multi-task AIS~\cite{nguyen2018multi}, which integrates recurrent neural networks with latent-variable modeling to simultaneously address trajectory reconstruction, anomaly detection, and vessel-type identification. PG-DPM~\cite{zhang2024long} introduces a physics-guided diffusion probabilistic model specifically designed for long-term vessel trajectory imputation by incorporating maritime domain knowledge into the generative process. 
Complementary studies on trajectory compression and clustering~\cite{li2020compression, li2021trace, li2022evolutionary, song2024quantifying} share similar goals of preserving motion patterns and data utility. 
However, these methods generally focus exclusively on reconstructing vessel positions and rarely consider other heterogeneous attributes, which are critical for comprehensive AIS data analysis.

\begin{table*}[!htbp]
  \centering
  \vspace{-2mm}
  \caption{Summary of AIS attributes categorized by type and time scale.}
  \label{tab:attribute_characteristics}
  \vspace{-3mm}
  \setlength{\tabcolsep}{2pt}\small
  \begin{tabular}{|l|c|c|c|c|c|c|c|c|c|}
  \hline
  \multirow{2}{*}{\textbf{Attribute (Symbol)}} & 
  \multicolumn{4}{c|}{\textbf{Attribute Type}} & 
  \multicolumn{5}{c|}{\textbf{Time Scale}} \\
  \cline{2-10}
  & \makecell{Spatio-temporal ($\mathbf{X}_s$)} & \makecell{Cyclical ($\mathbf{X}_c$)} & \makecell{Continuous ($\mathbf{X}_n$)} & \makecell{Discrete ($\mathbf{X}_d$)} & \makecell{S1 ($\mathbf{X}^{1}$)} & \makecell{S2 ($\mathbf{X}^{2}$)} & \makecell{S3 ($\mathbf{X}^{3}$)} & \makecell{S4 ($\mathbf{X}^{4}$)} & \makecell{S5 ($\mathbf{X}^{5}$)} \\
  \hline
  Longitude ($\lambda$) & \checkmark & & & & \checkmark & & & & \\
  \cline{1-10}
  Latitude ($\phi$) & \checkmark & & & & \checkmark & & & & \\
  \cline{1-10}
  Timestamp ($\tau$) & \checkmark & & & & \checkmark & & & & \\
  \hline
  True heading angle ($\psi$) & & \checkmark & & & & \checkmark & & & \\
  \cline{1-10}
  Course Over Ground ($\theta$) & & \checkmark & & & & \checkmark & & & \\
  \cline{1-10}
  Speed Over Ground ($s$) & & & \checkmark & & & \checkmark & & & \\
  \hline
  Navigation status ($\eta$) & & & & \checkmark & & & \checkmark & & \\
  \cline{1-10}
  Hazardous cargo type ($\chi$) & & & & \checkmark & & & & \checkmark & \\
  \cline{1-10}
  draught ($d$) & & & \checkmark & & & & & \checkmark & \\
  \hline
  Length ($\ell$) & & & \checkmark & & & & & & \checkmark \\
  \cline{1-10}
  Width ($\beta$) & & & \checkmark & & & & & & \checkmark \\
  \cline{1-10}
  Vessel type ($\kappa$) & & & & \checkmark & & & & & \checkmark \\
  \hline
  \multicolumn{10}{l}{$^{\dagger}$Si represents time scale $i$, where $i \in \{1,2,3,4,5\}$}\\
  \multicolumn{10}{l}{$^\ddagger$ Attribute types and time scales are organized according to Definition~\ref{def:attribute_characteristics} and \ref{def:multiple_time_scales}.}
  \end{tabular}
  \vspace{-2mm}
  \label{tab:ais_info}
\end{table*}
\section{Preliminaries}
\label{sec:preliminaries}
\subsection{Data and Notation}
\label{sec:DataNotation}
\begin{definition}[\textbf{AIS Record}]\label{def:ais_record} 
As summarized in Table~\ref{tab:ais_info}, an AIS Record $\mathcal{X}$ with $N$ attributes can be characterized from two perspectives: 
(i) \textbf{attribute type:} including spatio-temporal set $\mathbf{X}_s = \{\lambda, \phi, \tau\}$, cyclical set $\mathbf{X}_c = \{\psi, \theta\}$, continuous set $\mathbf{X}_n = \{s, d, \ell, \beta\}$, or discrete set $\mathbf{X}_d = \{\eta, \chi, \kappa\}$; (ii) \textbf{time scale (update rate):} attributes are grouped according to decreasing update rates into five time scales (detailed in Definition~\ref{def:multiple_time_scales}): $\mathbf{X}^{1} = \{\lambda, \phi, \tau\}$, $\mathbf{X}^{2} = \{\psi, \theta, s\}$, $\mathbf{X}^{3} = \{\eta\}$, $\mathbf{X}^{4} = \{\chi, d\}$, and $\mathbf{X}^{5} = \{\ell, \beta, \kappa\}$.
\end{definition} 

An attribute can be denoted by its attribute type and time scale. For instance, Speed Over Ground (SOG) is a continuous attribute $x_n^2$ at time scale 2, denoted by $s$. It indicates the vessel's speed, and simultaneously belongs to $\mathbf{X}^2$, $\mathbf{X}_n$, and $\mathbf{X}_n^2$.

\begin{definition}[\textbf{Vessel-specific AIS Record Sequence}]
A vessel-specific AIS record sequence is a heterogeneous multi-variable time series denoted by $\mathbf{X} = \langle \mathbf{X}_{1}, \mathbf{X}_{2}, \cdots, \mathbf{X}_{T}\rangle \in \mathbb{R}^{T \times N}$. Each record $\mathbf{X}_t$ at time step $t$ can be decomposed into two complementary views: by time scale (update rate), $\mathbf{X}_t = \{\mathbf{X}_t^{1}, \mathbf{X}_t^{2}, \mathbf{X}_t^{3}, \mathbf{X}_t^{4}, \mathbf{X}_t^{5}\}$; by attribute type, $\mathbf{X}_t = \{\mathbf{X}_{t,s}, \mathbf{X}_{t,c}, \mathbf{X}_{t,n}, \mathbf{X}_{t,d}\}$.
\end{definition}

For example, SOG $s_t$ at time step $t$ is a continuous attribute $x_{n, t}^2$ at time scale 2, belonging to $\mathbf{X}_{t, n}^{2}$, $\mathbf{X}_{t, n}$, and $\mathbf{X}_{t}^{2}$.

AIS data attributes can generally be categorized into three types based on their intrinsic characteristics: spatio-temporal, continuous, and discrete. 
To achieve a finer granularity and better applicability, we introduce a fourth category by extracting COG $\theta$ and THA $\psi$ from the continuous attributes into cyclical attributes due to their periodic nature. 
The final categorization is as follows:

\begin{definition}[\textbf{Intrinsic Characteristics of AIS Attributes}]
\label{def:attribute_characteristics}
AIS attributes can be categorized into four types based on their intrinsic characteristics, as summarized in Table~\ref{tab:ais_info}:
\begin{itemize}[itemsep=1pt, leftmargin=10pt, parsep=-1pt] 
\item \textbf{Spatio-temporal} attributes exhibit spatial constraints (e.g., latitude-dependent distortion and spherical continuity) and nested periodicities (e.g., daily, weekly cycles). 
\item \textbf{Cyclical} attributes recur periodically and exhibit smooth wrap-around at boundaries. 
\item \textbf{Continuous} attributes exhibit wide variations in scale, requiring appropriate normalization. 
\item \textbf{Discrete} attributes belong to finite categorical sets without inherent numerical ordering. 
\end{itemize}
\end{definition}

\textbf{Spatio-temporal attributes} exhibit spatial and temporal characteristics. For instance, spatial scaling varies with latitude: at the equator (\(\phi = 0^\circ\)), a longitudinal change of \(1^\circ\) corresponds to approximately 111\,km, while at a latitude of \(\phi = 79^\circ\) North, the same \(1^\circ\) shift corresponds to roughly 20\,km. Moreover, longitude values of \(\lambda = 180^\circ\) and \(\lambda = -180^\circ\) represent the same geographic meridian, due to spherical continuity. Temporally, timestamps demonstrate nested periodicities --- both daily (\(24 \times 3600\) seconds) and weekly (\(168 \times 3600\) seconds) cycles return to the starting point (\(\tau = 0\)).

\textbf{Cyclical attributes}, such as heading angle (\(\psi\)), measured from \(0^\circ\) to \(359^\circ\), demonstrates circular continuity, where angles exceeding \(359^\circ\) wrap around to \(0^\circ\) without causing discontinuity.

\textbf{Continuous attributes}, such as draught (\(d\)) and speed over ground (\(s\)), differ substantially in their numerical ranges. Draught typically ranges from \(0\) to \(20\) meters, while speed ranges from \(0\) to \(30\) knots or higher. Due to the distinct scales, a unit change in draught and a unit change in speed are not directly comparable, thus requiring normalization prior to unified analyses.

\textbf{Discrete attributes} represent categorical values without inherent numerical order. For example, the navigation status attribute (\(\eta\)) includes distinct categories such as \(0\) (under way), \(1\) (anchored), and \(2\) (not under command). These values function purely as categorical labels rather than as numerical measurements.
\subsection{Multi-scale Heterogeneous Graph}
AIS data attributes exhibit distinct update patterns that naturally lead to five time scales. According to the update mechanism~\cite{Mekkaoui_automatic_2022}, attributes can initially be categorized into three groups: (i) Autonomous sensor updates for the first six attributes in Table~\ref{tab:ais_info}; (ii) Periodic crew updates for the seventh and eighth attributes; (iii) One-time vessel registration updates for the last four attributes.

To capture finer temporal granularities, we further subdivide these groups: autonomous sensor attributes are split based on whether they require measurement intervals, while crew-updated attributes are divided based on whether they only change between voyages. This results in five distinct time scales that reflect the natural update frequencies of AIS attributes.

\begin{definition}[\textbf{Multiple Time Scales}] \label{def:multiple_time_scales} 
We categorize AIS data attributes into five distinct time scales based on their update rates, as summarized in Table~\ref{tab:ais_info}. 
Specifically:
\begin{itemize}[itemsep=1pt, leftmargin=10pt, parsep=-1pt] 
\item \textbf{Time scale 1}: Attributes $x^{1} \in \mathbf{X}^{1}$ have instantaneous values without requiring measurement intervals; thus, their theoretical update interval lower bound is zero. 
\item \textbf{Time scale 2}: Attributes $x^{2} \in \mathbf{X}^{2}$ change rapidly and require short measurement intervals, resulting in a theoretical lower bound slightly above zero. 
\item \textbf{Time scale 3}: Attributes $x^{3} \in \mathbf{X}^{3}$ change discretely at a moderate frequency during a voyage; hence, their update intervals have a theoretical lower bound above that of time scale 2. 
\item \textbf{Time scale 4}: Attributes $x^{4} \in \mathbf{X}^{4}$ remain constant within individual voyages but may change between voyages, setting their theoretical lower bound to the duration of the shortest voyage. 
\item \textbf{Time scale 5}: Attributes $x^{5} \in \mathbf{X}^{5}$ change rarely during a vessel’s operational lifespan, resulting in a very large theoretical lower bound for their update interval. 
\end{itemize}
\end{definition}

For example, SOG is classified as time scale 2, as it requires short intervals to accurately measure its rapidly changing value.

\begin{definition}[\textbf{Multi-scale Temporal Features}] 
AIS data attributes generate hierarchical features across different time scales. Specifically, for an attribute $x^{j} \in \mathbf{X}^{j}$ at its original time scale $j$, features $v_{x^{j}}^k \in \mathcal{V}^{k}$ can be derived at time scales $k$, where $k \geq j$. Thus, each attribute $x^{j}$ generates $5 + 1 - j$ multi-scale features that capture its temporal features at different time scales. 
\end{definition}

\begin{figure*}[!h]
    \centering
    \includegraphics[width=0.9\textwidth]{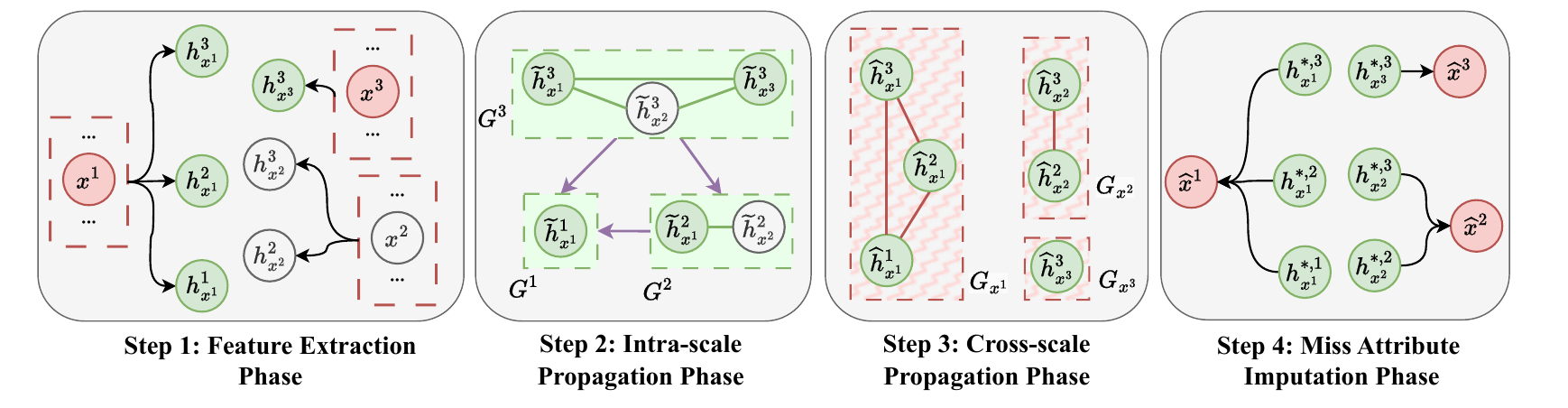}
    \caption{\warning{An example of missing attribute imputation based on a multi-scale heterogeneous graph.}}
    \label{fig:Example-MTS-HGNN}
\end{figure*}

\begin{figure}[!tbp]
    \centering
    \setlength{\abovecaptionskip}{0.00cm}
    \includegraphics[width=0.43\textwidth]{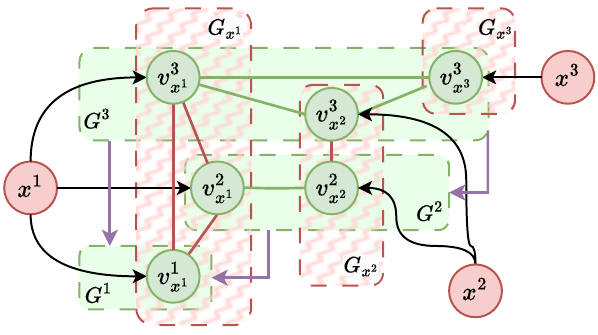}
    \caption{\warning{An example of multi-scale heterogeneous graph.}}
    \vspace{-4mm}
    \label{fig:heterogeneous_graph}
\end{figure}

\begin{example} \label{ex:heterogeneous_graph_1}
Consider three attributes $x^1$, $x^2$, and $x^3$ at time scales 1, 2, and 3, respectively (red nodes in Figure~\ref{fig:heterogeneous_graph}). Each attribute generates temporal features at its own and higher time scales: $x^1$ generates features $(v_{x^1}^1, v_{x^1}^2, v_{x^1}^3)$, $x^2$ generates features $(v_{x^2}^2, v_{x^2}^3)$, and $x^3$ generates the feature $v_{x^3}^3$ (green nodes in Figure~\ref{fig:heterogeneous_graph}).
\end{example}

\begin{definition}[\textbf{Multi-scale Heterogeneous Graph}]
\label{def:MHGraph}
A multi-scale heterogeneous graph $\mathcal{G} = (\mathcal{V}, \mathcal{E})$ represents temporal features derived from all  attributes across multiple time scales. Specifically, the node set $\mathcal{V}$ comprises the temporal features, while the edge set $\mathcal{E}$ captures dependencies between them. The graph structure integrates two types of relationships:
(i) \textbf{Time scale subgraphs} $\{\mathcal{G}^k\}_{k=1}^5$: each subgraph $\mathcal{G}^k=(\mathcal{V}^k, \mathcal{E}^k)$ is fully connected, consisting of nodes $\mathcal{V}^k$ denoting features at the same time scale $k$ as edges $\mathcal{E}^k$, capturing intra-scale dependencies; edges $\mathcal{E}^k$ are informed by higher-scale subgraphs $\mathcal{G}^{k+1}, \ldots, \mathcal{G}^5$, ensuring that relationships at lower time scales are contextually modulated by high-scale features;
(ii) \textbf{Attribute subgraphs} $\{\mathcal{G}_{x}\}_{x \in \mathbf{x}}$: each subgraph $\mathcal{G}_{x}=(\mathcal{V}_{x}, \mathcal{E}_{x})$ is fully connected, containing nodes denoting features of a specific attribute $x$ across multiple time scales. 
\warning{These attribute subgraphs are mutually independent, and} edges $\mathcal{E}_{x}$ capture cross-scale dependencies within the same attribute. 
\end{definition}

\begin{example}
Continuing Example~\ref{ex:heterogeneous_graph_1}, feature $v_{x^1}^2$ establishes intra-scale connections with $v_{x^2}^2$ through subgraph $\mathcal{G}^2$ (shown in the green box labeled $\mathcal{G}^2$), and the edge between $v_{x^1}^2$ and $v_{x^2}^2$ is modulated by higher-scale subgraphs $\mathcal{G}^3$ (shown in purple lines from $\mathcal{G}^3$ to $\mathcal{G}^2$).
Simultaneously, $v_{x^1}^2$ connects across scales within its attribute subgraph $\mathcal{G}{x^1}$ (shown as the red box labeled $\mathcal{G}{x^1}$), linking features $v_{x^1}^1$, $v_{x^1}^2$, and $v_{x^1}^3$. However, $v_{x^1}^2$ does not connect to $v_{x^2}^3$ due to differences in both attribute type and time scale. 
\end{example}

\subsection{Problem Formulation}
\label{sec:problem_formulation}
\begin{definition}[\textbf{Observation Mask}] 
Observation availability is represented by a binary mask $\boldsymbol{M} \in \{0,1\}^{T \times N}$, where $m_{x_t}=1$ indicates that attribute $x$ is observed at time step $t$, and $m_{x_t}=0$ indicates that the attribute is missing. For training and evaluation purposes, we select imputation targets $\widetilde{\boldsymbol{X}} \in \mathbb{R}^{T \times N}$ manually from the observed data and mark these selected positions using mask $\widetilde{\boldsymbol{M}} \in \mathbb{R}^{T \times N}$. 
\end{definition}


The masking pattern of $\widetilde{\boldsymbol{M}}$ varies across the attributes at different time scales in a record sequence: attributes $\mathbf{X}^5$ at time scale 5 generally exhibit either complete availability or are missing extensively; attributes $\mathbf{X}^4$ at time scale 4 present inter-voyage missing patterns, changing values mainly between voyages; attributes $\mathbf{X}^3$ at time scale 3 experience missing data related specifically to voyage phases; attributes $\mathbf{X}^2$ and $\mathbf{X}^1$ at time scales 1 and 2 frequently suffer from irregular and intermittent missing values due to signal interference or communication issues. 
\warning{The details of the masking pattern are discussed in Section \ref{sec:masking_strategies}.}

\noindent \begin{definition}[\textbf{Multi-scale Heterogeneous Graph-based Missing Attribute Imputation}]
\label{def:imputation_process}
The imputation process is formulated as a function $\mathcal{F}$ that maps a record sequence $\mathbf{X} * \mathbf{M}$ with missing attributes to a complete record sequence $\widetilde{\mathbf{X}}$:
\begin{equation}
\widetilde{\mathbf{X}} = \mathcal{F}(\mathbf{X} * \mathbf{M}),
\end{equation}
\warning{where $\widetilde{\mathbf{X}}$ is the imputed record sequence. Specifically, $\mathcal{F}(\cdot)$ can be divided into four phases:}
\setlength{\parsep}{0pt}
\begin{itemize}[itemsep=1pt, leftmargin=10pt]
    \item \textbf{Feature Extraction Phase} (\warning{$\mathbf{H} \leftarrow \mathcal{F}_1(\mathbf{X} * \mathbf{M})$}): \warning{Extract temporal representations $\mathbf{H}$ of all attributes at distinct scales from record sequence $\mathbf{X} * \mathbf{M}$ with missing attributes.}
    \item \textbf{Intra-scale Propagation Phase} (\warning{$\widetilde{\mathbf{H}} \leftarrow \mathcal{F}_2(\mathbf{H})$}): Propagate representations $\widetilde{\mathbf{H}}$ within time scale subgraphs $\{\mathcal{G}^k\}_{k=1}^5$ to ensure temporal consistency within each scale. 
    In this process, the edges $\{\mathcal{E}^k\}_{k=1}^5$ within each time scale subgraph is modulated by the higher-scale subgraphs, ensuring that the relationships at lower time scales are contextually modulated by high-scale features.
    \item \textbf{Cross-scale Propagation Phase} (\warning{$\hat{\mathbf{H}} \leftarrow \mathcal{F}_3(\widetilde{\mathbf{H}})$}): Propagate representations $\hat{\mathbf{H}}$ across different time scales within attribute subgraphs $\{\mathcal{G}_{x}\}_{x \in \mathcal{X}}$ to impute missing features.
    \item \textbf{Missing Attribute Imputation Phase} (\warning{$\widetilde{\mathbf{X}} \leftarrow \mathcal{F}_4(\mathbf{H}, \widetilde{\mathbf{H}}, \hat{\mathbf{H}})$}): Recover missing attributes $\widetilde{\mathbf{X}}$ based on $\mathbf{H}^{*}$ that combines the original representations $\mathbf{H}$ and propagated representations $\widetilde{\mathbf{H}}$ and $\hat{\mathbf{H}}$ from intra-scale and cross-scale propagation.
    \end{itemize}
\end{definition}


\begin{figure*}[ht]
    \centering
    \includegraphics[width=1\textwidth]{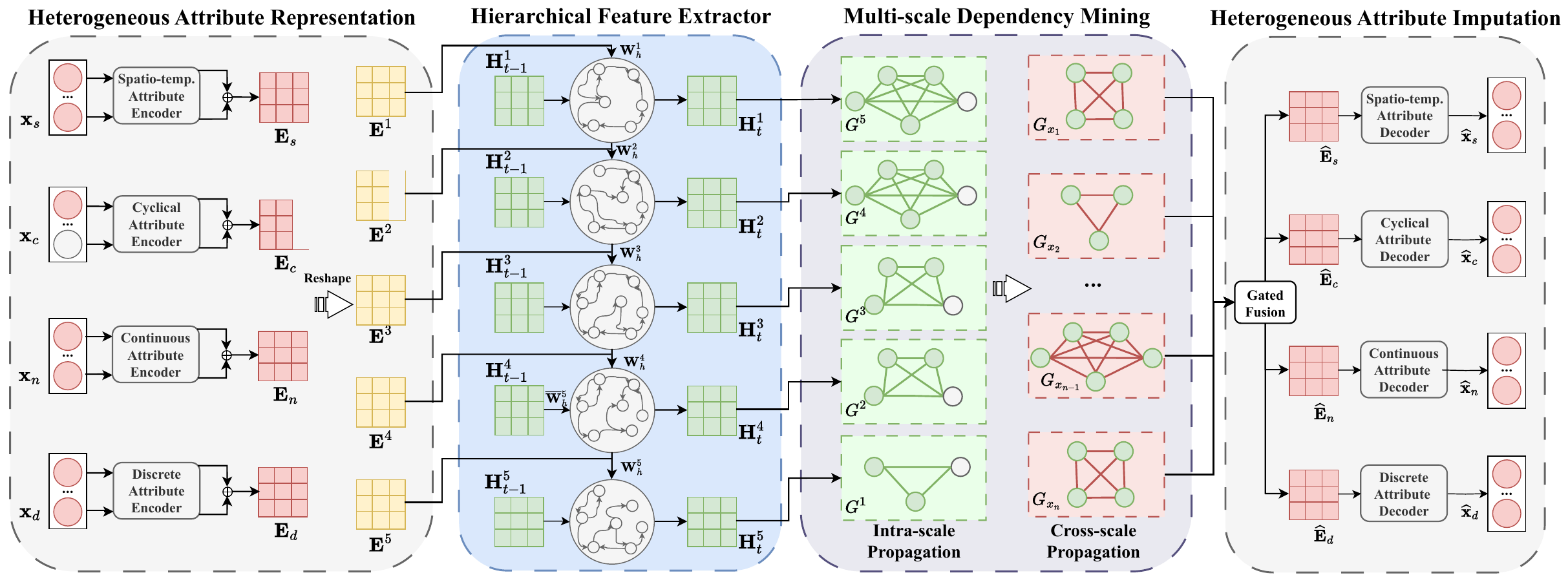}
    \caption{Overview of the multi-scale heterogeneous graph imputation network for AIS data.}
    \label{fig:mts-hg-overview}
\end{figure*}

\begin{example}\label{ex:imputation_process}
Continuing Example~\ref{ex:heterogeneous_graph_1}, as seen in Figure~\ref{fig:Example-MTS-HGNN}, the imputation for the missing attribute $x^2$ begins with multi-scale representation extraction.
Attributes at different scales produce representations: attribute $x^1$ generates representations ${h_{x^1}^1, h_{x^1}^2, h_{x^1}^3}$ (scales 1--3), attribute $x^3$ produces ${h_{x^3}^3}$ (scale 3), while attribute $x^2$ yields incomplete representations ${h_{x^2}^2, h_{x^2}^3}$ (scales 2--3).
Temporal consistency is maintained through intra-scale propagation on subgraphs $\mathcal{G}^1$, $\mathcal{G}^2$, and $\mathcal{G}^3$ (see the green box). 
The intra-scale propagation employs dynamic relationship modeling where edge weights in lower-scale subgraphs (e.g., $\mathcal{G}^2$) are dynamically modulated by features from higher-scale subgraphs (e.g., $\mathcal{G}^3$), enabling contextual adaptation of attribute relationships across different temporal scales.
This results in refined representations $\{\widetilde{h}_{x^1}^k\}_{k=1}^{3}$, $\{\widetilde{h}_{x^3}^k\}_{k=3}^{3}$, and partially recovered representations $\{\widetilde{h}_{x^2}^k\}_{k=2}^{3}$. 
Subsequent cross-scale propagation within attribute subgraphs $\mathcal{G}_{x^1}$, $\mathcal{G}_{x^2}$, and $\mathcal{G}_{x^3}$ (see the red box) synthesizes missing representations $\hat{h}_{x^2}^2$ and $\hat{h}_{x^2}^3$ through hierarchical representation fusion. 
The final imputation integrates original, intra-scale, and cross-scale representations as $h^{*,k}_{x^2} = [h_{x^2}^k, \widetilde{h}_{x^2}^k, \hat{h}_{x^2}^k]$ for $k\in\{1,2\}$, ultimately reconstructing the attribute $\hat{x}^2$ via a nonlinear transformation $\mathcal{F}(h^{*,1}_{x^2}, h^{*,2}_{x^2})$.
\end{example}

\section{Multi-scale Heterogeneous Graph-based Imputation Network} \label{sec:methodology}

Based on the four-phase imputation process (see Definition~\ref{def:imputation_process}), \textsf{MH-GIN} adopts an architecture comprising four core components, as illustrated in Figure~\ref{fig:mts-hg-overview}.
Initially, the framework extracts multi-scale temporal representations using the Heterogeneous Attribute Encoder and the Hierarchical Temporal Feature Extractor. These two components collaboratively capture hierarchical temporal features from each attribute while preserving their intrinsic attribute type characteristics. 
Next, the Multi-scale Dependency Mining Module constructs a multi-scale heterogeneous graph and performs missing feature imputation via a two-stage graph propagation process. Lastly, the Heterogeneous Attribute Imputation Module reconstructs missing attributes using specialized decoders tailored to each attribute type. For clarity and architectural consistency, we denote the dimensionality of hidden layers by \(d\), noting that the actual dimensionality may vary based on the requirements of individual components.

\subsection{Heterogeneous Attribute Encoder}
\label{sec:encoding-module}
As illustrated to the left in Figure~\ref{fig:mts-hg-overview}, this module aims to provide representations for different attribute types (as defined in Definition~\ref{def:attribute_characteristics}) while preserving their intrinsic characteristics. 

\noindent\textbf{Spatio-temporal Attribute Encoder.}
Spherical coordinates (longitude $\lambda$, latitude $\phi$, as defined in Definition~\ref{def:ais_record} and Table~\ref{tab:attribute_characteristics}) 
exhibit non-Euclidean geometry, as standard numeric encodings fails to capture spherical continuity and latitude-dependent distortion \cite{DBLP:conf/iclr/CohenGKW18}. We address this by extending 3D Cartesian projections \cite{wiki_cartesian_2023} with additional harmonic terms to enhance directional sensitivity. We then apply learnable affine transformations with hyperbolic tangent activations to produce embeddings $\mathbf{e}_{\lambda}$ and $\mathbf{e}_{\phi}$. 

Timestamp $\tau$ has nested periodicities (daily, weekly, monthly, yearly cycles). To preserve these temporal patterns for neural processing, we employ multi-frequency sinusoidal encoding that captures key temporal cycles \cite{10.24963/ijcai.2024/803}. The sinusoidal outputs then pass through a linear layer with the $\tanh$ activation function, producing a \(d\)-dimensional representation $\mathbf{e}_{\tau} \in\mathbb{R}^d$.

\noindent \textbf{Cyclical Attribute Encoder.}
Cyclical attributes \warning{$x_c \in \mathbf{X}_c = \{\psi, \theta\}$} exhibit boundary continuity requirements. To preserve rotational equivalence, we implement trigonometric encoding using sine and cosine transformations, followed by learnable linear layers with $\tanh$ activation. This approach maintains angular continuity in the embedding space while ensuring smooth transitions between neighboring angular values.

\noindent\textbf{Continuous Attribute Encoder.}
Continuous attributes \warning{$x_n \in \mathbf{X}_n = \{s, d, \ell, \beta\}$} range widely in scale, which can skew training dynamics. We address this via adaptive normalization using running statistics from the training set, with learnable scaling parameters for each attribute. A linear layer with ReLU activation then produces the final embeddings $\mathbf{e}_n$, helping preserve original distributions while stabilizing gradients.

\noindent\textbf{Discrete Attribute Encoder.}
Discrete attributes \warning{$x_d \in \mathbf{X}_d = \{\eta, \chi, \kappa\}$} represent categorical semantics without intrinsic ordering. We convert them to one-hot vectors and project through learnable transformations with $\tanh$ activation to obtain embeddings $\mathbf{e}_d$. 

Finally, all type-specific representations are concatenated into a unified embedding matrix $\mathbf{E}_t \in \mathbb{R}^{N \times d}$, preserving each attribute's intrinsic properties within a shared embedding space. 
\subsection{Hierarchical Temporal Feature Extractor}
\label{sec:Multiple Time-Scale  Mining Module}
As illustrated in the middle-left part of Figure~\ref{fig:mts-hg-overview}, this module captures multiple time-scale features via three complementary mechanisms: 1) Progressive temporal abstraction through increasing layer depth for finer time scales; 2) High-frequency smoothing via leaky integration in upper layers; 3) Fixed reservoir weights for computational efficiency. 
Concretely, we implement Deep Echo State Networks (DeepESN)~\cite{gallicchio_deep_2020} with leaky integrator neurons~\cite{JAEGER2007335} as our core architecture. 
Compared with other sequence models (LSTM~\cite{hochreiter1997long}, Transformer~\cite{vaswani2017attention}, etc.), their hierarchical reservoir structure naturally produces multi-scale representations through layer-wise abstraction, while fixed random weights enable efficient processing of AIS data. Moreover, the leaky integration mechanism provides explicit control over temporal memory retention, aligning with our time-scale hierarchy in Definition~\ref{def:multiple_time_scales}.

For an attribute \(x^k\) at time scale \(k \in \{1,\ldots,5\}\), we use \(5 + 1 -k\) recurrent layers, so that higher time-scale attributes require fewer layers. The state update is governed by:
\begin{equation}
\label{eq:desn}
\begin{aligned}
& \qquad \qquad \mathbf{h}_{t}^{k-1} = \mathbf{e}_{t}^{k},\\
\mathbf{\bar{h}}_{t}^{l} &= \tanh \Bigl(\mathbf{W}_{h}^{l} \cdot \mathbf{h}_{t}^{l-1}
+ \mathbf{\bar{W}}_{h}^{(l)} \cdot \mathbf{h}_{t-1}^{l}
+ \mathbf{b}_h^{(l)}\Bigr),\\
& \quad \mathbf{h}_{t}^{l} = (1-\gamma_l)\,\mathbf{h}_{t-1}^{l} + \gamma_l\,\mathbf{\bar{h}}_{t}^{l}, \\
\end{aligned}
\end{equation}
where \(\gamma_l \in (0,1]\) is the leak rate controlling temporal retention at layer~\(l\). Random weight matrices \(\mathbf{W}_{h}^{l}, \mathbf{\bar{W}}_{h}^{l} \in \mathbb{R}^{d \times d}\) and bias \(\mathbf{b}_h^{l} \in \mathbb{R}^{d}\) preserve the echo state property while inducing layer-specific dynamics. 
Attributes at time scale \(k\) are processed exclusively by layers \(l \in \{k,\ldots,5\}\), since lower-scale features (\(l < k\)) do not originate from higher-scale attributes (see Example~\ref{ex:heterogeneous_graph_1}). 
Equation~\ref{eq:desn} represents the forward-only version of DeepESN that enables sequential processing of streaming AIS data for real-time imputation. This formulation allows the model to impute missing values based solely on past and current information, making it suitable for safety-critical maritime applications requiring immediate response. In offline scenarios where complete trajectory data is available, we can replace the forward-only DeepESN with a \textbf{bidirectional version} to achieve enhanced performance. The default setting of MH-GIN is the bidirectional version.


Finally, we denote the consolidated multi-scale temporal representation at time step \(t\) by:
\begin{equation}
\label{eq:H_final}
\mathbf{H}_{t} 
= \bigl[\mathbf{H}_{t}^{1}, \,\ldots,\, \mathbf{H}_{t}^{5}\bigr],    
\end{equation}
where \(\mathbf{H}_{t}^{l} = \bigl[\mathbf{h}_{t,1}^{l}, \ldots, \mathbf{h}_{t,N_l}^{l}\bigr]\) aggregates temporal features at scale \(l\), with \(N_l\) denoting both the feature count at scale \(l\) and the number of attributes from higher or equal scales (\(k \geq l\)).
\subsection{Multi-scale Dependency Mining Module}
As illustrated in the middle-right part of Figure~\ref{fig:mts-hg-overview}, the Multi-scale Dependency Mining Module performs three main functions: 1) aligning features within same time scale for temporal consistency, 2) learning dependencies between features in partially observed data, and 3) recovering missing features from incomplete attribute sets.
Conventional GNNs (e.g. GAT \cite{DBLP:conf/iclr/VelickovicCCRLB18}) struggle under these conditions, as missing attributes contaminate adjacency estimation (through corrupted node similarity measures) and cause cascading errors in feature updates.
Specifically, we uses a two-stage propagation mechanism to capture time-varying dependencies between features at different time scales.

\noindent\textbf{Stage 1: Intra-scale Propagation.} This phase aligns features that occur at the same time scale to ensure their temporal patterns are synchronized. 

To capture time-varying dependencies between different features at the same time scale, we model the edges within each time-scale subgraph $\mathcal{G}^k = (\mathcal{V}^k, \mathcal{E}^k)$ as dynamic rather than static. The weight of an edge is computed dynamically at each time step, explicitly modulated by corresponding features from higher time scales. Specifically, for each time scale subgraph with representations $\mathbf{H}^k_t$, we construct a dynamic adjacency matrix $\mathbf{\hat{A}}^k_t$ that captures time-varying dependencies between features at time scale $k$:

\begin{equation}
\mathbf{\hat{A}}^k = f^k_{\text{edge}}([\mathbf{H}^{k+1}; ... ;\mathbf{H}^5]) + \mathbf{A}^k,
\end{equation}
\warning{where $f^k_{\text{edge}}$ is a learnable function that computes edge weights based on both current scale features $\mathbf{H}^k$ and higher-scale contextual features $\mathbf{H}^{k+1}$ and $\mathbf{A}^k$ is a bias matrix. This allows edge weights at lower scales to be adjusted based on contextual features from higher scales, enabling the model to capture time-varying dependencies. The intra-scale propagation is then performed as:}
\begin{equation}
\mathbf{\tilde{H}}^k = (\mathbf{D}^k)^{-1/2} \cdot \mathbf{\hat{A}}^k \cdot (\mathbf{D}^k)^{-1/2} \cdot \mathbf{H}^k,
\end{equation}
\warning{where $\mathbf{D}^k = \text{diag}(\sum_j \mathbf{\hat{A}}^k_{ij})$ is the diagonal degree matrix that normalizes node degrees to prevent over-smoothing.}

\noindent\textbf{Stage 2: Cross-scale Propagation.} 
Building on temporally aligned features, this phase integrates multi-scale features through attribute subgraphs. 
We keep the edges within attribute subgraphs ($\mathcal{G}_x$) static since higher-scale features already provide hierarchical context. Dynamic cross-scale pathways would create recursive dependencies that introduce information redundancy and hinder convergence~\cite{tishby2015deep}. Specifically, for each attribute $x$ with multi-scale representations $\mathbf{\tilde{H}}_x$, we learn cross-scale interactions via:
\begin{equation}
\mathbf{\hat{H}}_{x} = (\mathbf{D}_x)^{-1/2} \cdot \mathbf{\hat{A}}_x \cdot (\mathbf{D}_x)^{-1/2} \cdot \mathbf{\tilde{H}}_x,
\end{equation}
where the learnable matrix $\mathbf{\hat{A}}_x$ models hierarchical temporal dependencies. The final representation combines both stages through residual connection:
\begin{equation}
\label{eq7}
\mathbf{h}^{*,k}_x = [\mathbf{h}_x^k; \mathbf{\tilde{h}}_x^k; \mathbf{\hat{h}}^k_x].
\end{equation}

\noindent\textbf{Theoretical Analysis.}
The multi-scale propagation mechanism ensures both numerical stability and robustness to input perturbations through its symmetric normalization design. We establish these properties through two key theoretical results.

\begin{lemma}[Stability of Multi-scale Propagation]
The multi-scale propagation mechanism is numerically stable if the spectral radius of each propagation matrix satisfies $\rho(\mathbf{P}) = \mathbf{D}^{-1/2} \hat{\mathbf{A}} \mathbf{D}^{-1/2} \leq 1$.
\end{lemma}

This lemma ensures that feature representations remain bounded throughout the propagation process, preventing numerical instability such as gradient explosion or vanishing. The symmetric normalization in both intra-scale and cross-scale propagation ensures that each stage acts as a non-expansive mapping, making the complete two-stage mechanism numerically stable (proof in Appendix~\ref{sec:appendix_stability_of_multi_scale_propagation}).

\begin{lemma}[Robustness of Multi-scale Propagation]
The multi-scale propagation operator $\mathcal{G}$ is Lipschitz continuous with bounded Lipschitz constant.
\end{lemma}

This property ensures that small perturbations to the input features (e.g., from missing data or noise) result in predictably bounded changes in the final output. The Lipschitz continuity arises from the composition of linear operators with bounded spectral norms, making the model robust to input uncertainties commonly encountered in real-world AIS data (proof in Appendix~\ref{sec:appendix_robustness_of_multi_scale_propagation}).

\color{black}
\subsection{Heterogeneous Attribute Imputation}
\label{sec:decoding-module}
As illustrated in the right part of Figure~\ref{fig:mts-hg-overview}, this module completes the imputation cycle by reconstructing missing attributes through specialized decoders for each attribute type. 

\noindent \textbf{Gated Fusion.}
The imputation process begins with a gated fusion mechanism that adaptively integrates information across multiple time scales, enabling the model to prioritize the most relevant temporal features for each missing attribute before applying type-specific reconstruction. Given the cross-scale representations $\mathbf{h}^{*}_{x^k}$, we apply gated fusion across time scales: 
\begin{equation}
\mathbf{g} = \sigma\left(\mathbf{W}_g  \cdot [\mathbf{h}^{*,1}_{x^k}; \ldots; \mathbf{h}^{*,5}_{x^k}] + \mathbf{b}_g\right),
\end{equation}

\begin{equation}
\label{eq:gated_fusion}
\mathbf{\tilde{e}}_{x^k} = \sum_{k=1}^5 g^{l} \odot (\mathbf{W}_e^{l} \cdot \mathbf{h}^{*,l}_{x^k}),
\end{equation}
where $\mathbf{g} \in [0,1]^5$ are learnable gating weights, $\odot$ denotes element-wise multiplication, $\mathbf{W}_g \in \mathbb{R}^{5\times 5d}$ aligns dimensions for scale integration, and $\mathbf{h}^{*,l}_{x^k}$ is set to zero vector when $l$ is lower than $k$.

\noindent \textbf{Spatio-temporal Decoder.} 
For coordinate attributes, we leverage the inherent spatial continuity of vessel trajectories by using local averages as base estimates, enabling the network to focus on learning incremental adjustments rather than absolute positions. Given missing coordinate attributes $\lambda_i$ and $\phi_i$ at time step $i$, we first compute the mean coordinates from nearby valid observations within a window $\mathcal{N}_i$. The model then computes spatial adjustments $\delta_{\lambda}$ and $\delta_{\phi}$ using the gated fusion embeddings $\mathbf{\tilde{e}}_{\lambda_i}, \mathbf{\tilde{e}}_{\phi_i}$ from Eq.~\ref{eq:gated_fusion}. The final imputed coordinates are obtained as $\hat{\lambda} = \lambda_{\text{base}} + \delta_{\lambda}$ and $\hat{\phi} = \phi_{\text{base}} + \delta_{\phi}$.

For timestamp attribute $\tau$, we predict time intervals between consecutive records rather than absolute timestamps, modeling temporal occurrences as a temporal point process \cite{lin2022exploring,omi2019fully}. We implement a conditional intensity function $\eta(\tau)$ using neural networks with Softplus activation to ensure non-negativity. The predicted timestamp is derived as $\hat{\tau}_{i} = \tau_{i-1} - \frac{\log(u)}{\eta(\tau)}$, where $u \sim \text{Uniform}(0, 1)$.

\noindent \textbf{Cyclical Decoder.} 
To preserve cyclical continuity, the cyclical decoder reconstructs angular values using trigonometric projection. The decoder first transforms the gated fusion embedding $\mathbf{\tilde{e}}_{x_c}$ through a two-layer neural network to generate intermediate representation $\mathbf{h}_c \in \mathbb{R}^2$. This intermediate representation is then normalized to unit length $\hat{\mathbf{e}}_{c} = \frac{\mathbf{h}_c}{\|\mathbf{h}_c\|_2}$ to ensure trigonometric identity $\hat{e}_{c,1}^2 + \hat{e}_{c,2}^2 = 1$. The final angular value is decoded as $\hat{x}_c = \frac{180}{\pi} \arctan_2(\hat{e}_{c,1}, \hat{e}_{c,2})$.

\noindent \textbf{Continuous Decoder.} 
For continuous attributes, we transform the gated fusion embedding $\mathbf{\tilde{e}}_{x_n}$ through a ReLU-activated linear layer to generate normalized representation $h_n$, then apply inverse normalization using the encoder's parameters $\alpha,\beta$ (see Section~\ref{sec:encoding-module}) and training statistics $\mu,\sigma$ to restore the original scale as $\hat{x}_n$.

\noindent\textbf{Discrete Decoder.} 
For discrete attributes, we apply a linear layer followed by softmax to predict class probabilities $\hat{y}$ from the gated fusion embedding $\mathbf{\tilde{e}}_{x_d}$, where the output dimension corresponds to the number of discrete classes $|C|$.

\subsection{Training}
To optimize \textsf{MH-GIN}, we design a comprehensive loss function that combines specialized reconstruction losses for each heterogeneous attribute type:
\begin{equation}
\mathcal{L}_{\textit{total}} = \lambda_{\textit{coo}} \mathcal{L}_{\textit{coo}} + \lambda_{\tau} \mathcal{L}_{\tau} + \lambda_{\textit{period}} \mathcal{L}_{\textit{period}} + \lambda_{\textit{cont}} \mathcal{L}_{\textit{cont}} + \lambda_{\textit{disc}} \mathcal{L}_{\textit{disc}},
\end{equation}
where $\lambda_{\textit{coo}}$, $\lambda_{\tau}$, $\lambda_{\textit{period}}$, $\lambda_{\textit{cont}}$, and $\lambda_{\textit{disc}}$ are balancing hyperparameters. The training employs specialized loss functions tailored to each attribute type: Haversine distance $\mathcal{L}_{\textit{coo}}$~\cite{wiki_haversine_2023} for spatio-temporal coordinates to capture spherical distance, mean squared error $\mathcal{L}_{\tau}$ for timestamp intervals, trigonometric encoding loss $\mathcal{L}_{\textit{period}}$ for cyclical attributes to preserve periodicity, mean squared error $\mathcal{L}_{\textit{cont}}$ for continuous attributes, and cross-entropy loss $\mathcal{L}_{\textit{disc}}$ for discrete classification. This multi-objective design ensures that each attribute type is reconstructed according to its inherent characteristics and constraints. 

\section{Experiments}
\label{sec:experiment}

\subsection{Overall Settings}
\subsubsection{Datasets.} 
We use two AIS datasets: AIS-DK from the Danish Maritime Authority\footnote{\url{http://aisdata.ais.dk/?prefix=2024/}(Last access: 2025.10)} and AIS-US from NOAA\footnote{\url{https://coast.noaa.gov/htdata/CMSP/AISDataHandler/2024}(Last access: 2025.10)}. \textbf{AIS-DK} covers Danish waters from March to December 2024, including major shipping routes in the Baltic and North Seas. It contains 18,481 vessel sequences and 10,518,249 AIS records, with an average sequence duration of 9 days. \textbf{AIS-US} spans March to May 2024, focusing on US coastal waters with dense traffic near major ports. It includes 19,091 vessel sequences and 10,546,297 AIS records, with an average sequence duration of 4 days. 

Both datasets feature diverse vessel types and often capture multiple voyages per sequence, especially for short-distance operations. Their high temporal resolution and broad coverage provide a robust basis for evaluating our methods. 


\subsubsection{Evaluation Metrics}
Our evaluation metrics are tailored to the distinct characteristics of AIS data attributes. For continuous numerical values, we employ Mean Absolute Error (MAE) and Symmetric Mean Absolute Percentage Error (SMAPE). For categorical attributes, we utilize Accuracy (ACC). 

\subsubsection{Baseline Methods}
To evaluate the performance of our proposed method, we compare with classic models and state-of-the-art methods for multi-variable time-series imputation, spatio-temporal imputation, and trajectory imputation. The baselines include: 
statistical methods (MEAN, KNN, Lin-ITP), 
classic models (MF, TRMF~\cite{yu2016temporal}), 
multi-variable time-series imputation (CSDI~\cite{tashiro2021csdi}), 
spatiotemporal imputation (ImputeFormer, PriSTI~\cite{liu_pristi_2023}), 
and trajectory imputation (Multi-task AIS~\cite{nguyen2018multi}, PG-DPM~\cite{zhang2024long}).

\subsubsection{Masking Strategies.}
\label{sec:masking_strategies}
AIS data presents distinctive missing patterns across various time scales (as discussed in Section \ref{sec:problem_formulation}). To accurately simulate these real-world scenarios, we implement three targeted masking strategies with mask ratio $r$:
\setlength{\parsep}{0pt}
\begin{itemize}[itemsep=2pt, leftmargin=10pt, parsep=-1pt]
\item \textbf{Point Masking} (Scales 1 and 2): Randomly masks individual values throughout the sequence with probability \(r\), simulating sporadic missingness characteristic of high-frequency attributes.
\item \textbf{Block Masking} (Scales 3 and 4): Masks continuous segments within individual voyages throughout the sequence with probability \(r\), capturing voyage-phase-related missing patterns common to these attributes.
\item \textbf{Entire Masking} (Scale 5): Masks entire attribute sequences for a proportion \(r\) of vessels, replicating the systematic absence of vessel-specific data frequently observed in AIS records.
\end{itemize}

\begin{table*}[!htbp]
    \small
    \setlength{\tabcolsep}{1.55pt}
    \caption{{Overall effectiveness comparison on AIS-DK.}}
    \vspace{-2mm}
    \label{tab:performance_comparison-AIS-DK}
    \begin{tabular}{l|c|cc|cc|cc|cc|cc|cc|cc|c}
    \hline
\multirow{3}{*}{\textbf{Method}} & \makecell[c]{Coord. $(\lambda,\phi)$} & \multicolumn{2}{c|}{\makecell[c]{Time $\tau$}} & \multicolumn{2}{c|}{\makecell[c]{Head. $\psi$}} & \multicolumn{2}{c|}{\makecell[c]{Course $\theta$}} & \multicolumn{2}{c|}{\makecell[c]{Speed $s$}} & \multicolumn{2}{c|}{\makecell[c]{Draft $d$}} & \multicolumn{2}{c|}{\makecell[c]{Length $\ell$}} & \multicolumn{2}{c|}{\makecell[c]{Width $\beta$}} & \makecell[c]{Type $\kappa$} \\
    \cmidrule{2-17}
    & Dist.     & MAE     & SMAPE  & MAE     & SMAPE  & MAE     & SMAPE  & MAE   & SMAPE & MAE    & SMAPE  & MAE     & SMAPE & MAE    & SMAPE & ACC     \\
    \midrule
    MEAN        & 1.313e-2 & 371.970 & 0.934  & 0.861   & 0.375  & 0.916   & 0.392  & 2.318 & 0.400 & 1.883 & 0.812 & -       & -     & -      & -     & -       \\
    KNN         & 8.432e-4 & 34.880  & 0.404  & 0.953   & 0.392  & 0.509   & 0.465  & 0.875 & 0.453 & \underline{1.310}  & 0.722  & -       & -     & -      & -     & -       \\
    Lin-ITP     & 5.294e-4 & \underline{33.243}  & \underline{0.396}  & 2.895   & 0.429  & 8.950   & 1.478  & 0.792 & 0.486 & 1.460  & 0.821  & -       & -     & -      & -     & -       \\
    MF          & 3.314e-1 & 57.133  & 0.728  & 147.172 & 1.420  & 148.621 & 1.421  & 6.955 & 0.963 & 4.399  & 0.957  & 189.646  & 1.628 & 29.426  & 1.734 & -    \\
    TRMF        & 1.341e-1 & 37.604  & 0.479  & 51.378  & 0.869  & 55.934  & 0.3972 & 1.580 & 0.381 & 1.968  & 0.839  & 62.843   & 0.333 & 9.308 & 0.325 & -  \\ 
    CSDI        & 1.583e+0 & 100.375 & 1.588  & 0.388   & 0.390  & 0.432   & 0.246  & 0.745 & 0.397 & 7.786  & 0.625  & 141.077 & 1.514 & 18.240 & 1.120 & -   \\
    PriSTI      & 1.310e+0 & 71.486  & 0.857  & \underline{0.289}   & \underline{0.364}  & \underline{0.286}   & \underline{0.172}  & \underline{0.716} & \underline{0.373} & 6.186  & \underline{0.399}  & 145.274 & 1.273 & 8.456  & 0.273 & -   \\
    ImputeFormer & 1.839e-1 & 48.375  & 0.689  & 0.486  & 0.434  & 0.776  & 0.386  & 1.344 & 0.492 & 8.713  & 0.782  & \underline{46.055}  & \underline{0.194}  & \underline{7.687}  & \underline{0.259}  & -       \\
    Multi-task AIS & 3.863e-4 & -       & -      & -       & -      & -       & -      & -     & -     & -      & -      & -       & -     & -      & -     & \underline{0.322}     \\
    PG-DPM      & \underline{1.777e-4} & -       & -      & -       & -      & -       & -      & -     & -     & -      & -      & -       & -     & -      & -     & -       \\
    \midrule
    \textsf{MH-GIN} & \warninge{\textbf{1.112e-4}} & \warninge{\textbf{21.413}} & \warninge{\textbf{0.127}} & \warninge{\textbf{0.049}} & \warninge{\textbf{0.036}} & \warninge{\textbf{0.141}} & \warninge{\textbf{0.071}} & \warninge{\textbf{0.198}} & \warninge{\textbf{0.252}} & \warninge{\textbf{1.017}} & \warninge{\textbf{0.126}} & \warninge{\textbf{34.192}} & \warninge{\textbf{0.178}} & \warninge{\textbf{4.896}} & \warninge{\textbf{0.175}} & \warninge{\textbf{0.669}} \\
    \hline
    \end{tabular}
\end{table*}

\begin{table*}
    \small
    \caption{{Overall effectiveness comparison on AIS-US.}}
    \vspace{-2mm}
    \label{tab:performance_comparison-AIS-US}
    \setlength{\tabcolsep}{1.55pt}
    \begin{tabular}{l|c|cc|cc|cc|cc|cc|cc|cc|c}
    \hline
    \multirow{3}{*}{\textbf{Method}}  & \makecell[c]{Coord. $(\lambda,\phi)$} & \multicolumn{2}{c|}{\makecell[c]{Time $\tau$}} & \multicolumn{2}{c|}{\makecell[c]{Head. $\psi$}} & \multicolumn{2}{c|}{\makecell[c]{Course $\theta$}} & \multicolumn{2}{c|}{\makecell[c]{Speed $s$}} & \multicolumn{2}{c|}{\makecell[c]{Draft $d$}} & \multicolumn{2}{c|}{\makecell[c]{Length $\ell$}} & \multicolumn{2}{c|}{\makecell[c]{Width $\beta$}} & \makecell[c]{Type $\kappa$} \\
    \cmidrule{2-17}
    & Dist.     & MAE      & SMAPE   & MAE      & SMAPE   & MAE      & SMAPE   & MAE    & SMAPE  & MAE     & SMAPE   & MAE      & SMAPE  & MAE     & SMAPE  & ACC     \\
    \midrule
    MEAN        & 1.604e-2 & 260.317 & 0.491   & 0.908    & 0.370   & 1.203    & 0.449   & 2.433  & 1.344  & 3.164   & 0.476   & -        & -      & -       & -      & -       \\
    KNN         & 7.818e-4 & 44.358  & 0.516   & 0.452    & 0.391   & 0.846    & 0.640   & 0.702  & 1.020  & \underline{2.134}   & \underline{0.322}   & -        & -      & -       & -      & -       \\
    Lin-ITP     & 4.896e-4 & \underline{31.013}  & \underline{0.486}   & 7.617    & 0.365   & 28.444   & 0.487   & 0.687  & 0.968  & 2.429   & 0.394   & -        & -      & -       & -      & -       \\
    MF        & 6.830e-1 & 160.671 & 0.702   & 128.235  & 1.115   & 131.286  & 1.008   & 3.701  & 1.303  & 5.499   & 1.123   & 181.249  & 1.763  & 29.730  & 1.842  & -       \\
    TRMF          & 3.536e-1 & 232.299 & 0.945   & 68.246   & 0.491   & 82.993   & 0.580   & 1.489  & 1.279  & 2.617   & 0.462   & 58.320   & 0.366  & 8.326   & 0.307  & -       \\ 
    CSDI        & 1.552e+0 & 373.14  & 1.993   & 0.437   & 0.245   & 0.689   & 0.364   & 0.663  & 0.857   & 8.096   & 1.745   & 154.77   & 1.887  & 18.047  & 1.025  & -       \\
    PriSTI      & 1.663e+0 & 329.957 & 1.608   & \underline{0.348}    & \underline{0.233}   & \underline{0.595}  & \underline{0.329}   & \underline{0.591}  & \underline{0.787}  & 4.997   & 0.510   & 134.655  & 1.202  & 10.038  & 0.397  & -       \\ 
   ImputeFormer & 2.104e-1 & 110.320 & 0.536   & 1.138   & 0.570   & 1.216   & 0.505   & 1.865  & 0.976  & 4.360   & 0.943   & \underline{45.337}   & \underline{0.318}  & \underline{6.820}   & \underline{0.268}  & -       \\
 Multi-task AIS & 3.627e-4 & -       & -       & -        & -       & -        & -       & -      & -      & -       & -       & -        & -      & -       & -      & \underline{0.355}       \\
    PG-DPM      & \underline{1.512e-4} & -       & -       & -        & -       & -        & -       & -      & -      & -       & -       & -        & -      & -       & -      & -       \\
    \midrule
    \textsf{MH-GIN} & \warninge{\textbf{8.281e-5}} & \warninge{\textbf{15.146}} & \warninge{\textbf{0.016}} & \warninge{\textbf{0.104}} & \warninge{\textbf{0.072}} & \warninge{\textbf{0.461}} & \warninge{\textbf{0.184}} & \warninge{\textbf{0.368}} & \warninge{\textbf{1.182}} & \warninge{\textbf{1.899}} & \warninge{\textbf{0.314}} & \warninge{\textbf{38.263}} & \warninge{\textbf{0.273}} & \warninge{\textbf{5.382}} & \warninge{\textbf{0.213}} & \warninge{\textbf{0.446}} \\
    \hline
    \end{tabular}
\end{table*}

\subsubsection{\warning{Noise Injection Strategies.}}
\label{sec:noise_injection_strategies}
To evaluate the robustness of our proposed method, we inject noise into the AIS data to simulate real-world data corruption scenarios. We design a unified noise injection strategy controlled by parameter $\gamma$ that provides comparable corruption levels across all heterogeneous attribute types. For continuous attributes, we add Gaussian noise scaled by the attribute's own value $x'_{n} = x_{n} + \mathcal{N}(0, (\gamma x_n)^2)$.
For spatio-temporal coordinates, noise is scaled by movement dynamics to simulate GPS drift $\lambda' = \lambda + \mathcal{N}(0, (\gamma \sigma_{\Delta\lambda})^2)$ and $\phi' = \phi + \mathcal{N}(0, (\gamma \sigma_{\Delta\phi})^2)$. For timestamps, we corrupt time intervals rather than absolute values $\Delta\tau'_{i} = \Delta\tau_i + \max\left(0, \mathcal{N}(0, (\gamma \cdot \sigma_{\Delta\tau})^2)\right)$. For cyclical attributes like headings, we add angular noise with proper wrap-around handling $x'_{c} = \left( x_{c} + \mathcal{N}(0, (\gamma \cdot \sigma_{\Delta x_c})^2) \right) \pmod{360}$. For discrete categories, we implement label flipping with probability $\gamma$. This unified approach ensures that $\gamma$ serves as an interpretable control parameter for systematic noise injection across the entire heterogeneous dataset. 

\subsubsection{Experimental Settings}
All experiments are conducted on a server with Intel Xeon Processor (Icelake) CPUs, 100GB RAM, and two NVIDIA A10 GPUs (each with 23GB memory). For training our model, we use the Adam optimizer with a learning rate of $1e^{-3}$ and weight decay of $1e^{-4}$. The batch size is set to 64, and we train for 100 epochs. Early stopping with a patience of 10 epochs was applied to prevent overfitting. 
For data processing, we use an 80\%--10\%--10\% split for training, validation, and testing, respectively. 
For the masked value generation, we create synthetic missing data according to the three strategies described earlier, with mask ratios of 10\%, 20\%, 30\%, 40\% and 50\% to evaluate the robustness of different imputation methods. \warning{For the noise injection, we conduct experiments by varying the noise intensity $\gamma$ progressively from 0.0 to 0.5.} The default mask ratio is set to 30\% for all experiments unless otherwise specified. 
For methods, MEAN, Lin-ITP, and MF, no hyperparameters need to be set. We set $k$ of KNN to 20. For TRMF, CSDI, PriSTI, ImputeFormer, and Multi-task AIS, we use the default settings provided by the authors.
The more detailed experimental setting is provided in Appendix~\ref{app:detaledExpSetting}.
\subsection{Performance Evaluation}
\begin{table}[!tbp]
\centering
\small
\setlength{\tabcolsep}{3pt}
\caption{{Overall time cost and memory cost comparison.}}
\vspace{-2mm}
\label{tab:efficiency_comparison}
\renewcommand{\arraystretch}{1.1}
\begin{tabular}{l|c|c|c|c}
\toprule
\multirow{3}{*}{\textbf{Method}} & \multicolumn{2}{c|}{\textbf{AIS-DK}} & \multicolumn{2}{c}{\textbf{AIS-US}} \\
\cmidrule(lr){2-5}
& \textbf{Time (s)} & \textbf{Memory (MB)} & \textbf{Time (s)} & \textbf{Memory (MB)} \\
\midrule
MEAN                      & 1.24    & -       & 1.28    & -         \\
KNN                       & 5.39    & -       & 7.50    & -         \\
Lin-ITP                   & 1.26    & -       & 1.29    & -         \\
MF                        & 27.24   & -       & 28.29   & -         \\
TRMF                      & 34.02   & -       & 35.86   & -         \\
CSDI                      & 60.48   & 1.72    & 60.74   & 1.92      \\
PriSTI                    & 94.92   & 2.89    & 100.80  & 3.19      \\    
ImputeFormer              & 14.10   & 5.13    & 15.42   & 5.52      \\
Multi-task AIS            & 38.30   & 1.66    & 40.59   & 1.84      \\
PG-DPM                    & 120.46  & 1.79    & 128.32  & 1.99      \\
\midrule
\textsf{MH-GIN}           & \warninge{10.9}    & \warninge{1.43}    & \warninge{11.2}     & \warninge{1.83}     \\
\bottomrule
\end{tabular}
\vspace{-1mm}
\end{table}

\subsubsection{Effectiveness Analysis}
Tables \ref{tab:performance_comparison-AIS-DK} and \ref{tab:performance_comparison-AIS-US} present the performance comparison between \textsf{MH-GIN} and baseline methods. 
Navigation status $\eta$ and hazardous cargo type $\chi$ are excluded from Tables \ref{tab:performance_comparison-AIS-DK} and \ref{tab:performance_comparison-AIS-US} due to the lack of suitable baseline methods capable of imputing these specific categorical attributes. However, we emphasize that both categorical attributes are fully included in all subsequent experiments, including ablation studies, missing ratio analysis, and robustness evaluation.

\textsf{MH-GIN} consistently outperforms all baseline methods across both datasets and all attribute types, demonstrating its strong capability to leverage multi-scale dependencies between attributes for accurate AIS data imputation. Specifically, we observe the following: 
1) Spatio-temporal attributes — Coordinate $(\lambda, \phi)$ and Timestamp $\tau$: Lin-ITP achieves the best timestamp imputation performance by ensuring interpolated values fall within the temporal range of consecutive records. Similarly, PG-DPM performs well in coordinate imputation by incorporating physics-guided constraints on vessel movement. Although these methods are specifically designed for spatio-temporal attributes, \textsf{MH-GIN} still surpasses both with \warning{35.59\%--96.71\%} performance gains, which strongly demonstrates the effectiveness of modeling multi-scale dependencies. It also confirms that \textsf{MH-GIN}'s spatio-temporal encoder-decoder is not only effective in its own right, but also sufficiently expressive to support subsequent multi-scale dependency mining.
2) Cyclical attributes — Heading $\psi$ and Course $\theta$: Among baselines, PriSTI achieves the best results for cyclical attribute imputation. However, due to the dynamic nature of spatial information in AIS data, PriSTI and ImputeFormer fail to fully utilize spatial context, resulting in comparable performance to CSDI. While these methods outperform classic baselines, they struggle with boundary continuity in cyclical attributes. In contrast, \textsf{MH-GIN} achieves \warning{22.52\%--98.93\%} improvements over PriSTI, demonstrating the strength of its cyclical encoder-decoder in modeling cycical patterns while preserving boundary continuity.
3) Continuous attributes --- Speed $s$, Draft $d$, Length $\ell$, and Width $\beta$: \textsf{MH-GIN} yields \warning{2.48\%--72.35\%} improvements over PriSTI and ImputeFormer, primarily due to its ability to handle scale variation through a dedicated continuous encoder-decoder architecture.  Notably, CSDI, PriSTI and ImputeFormer generally assume that all variables are continuous attributes, which gives them a natural advantage when dealing with this attribute type. Despite this, \textsf{MH-GIN} still surpasses their performance, highlighting the importance and effectiveness of mining multi-scale dependencies across heterogeneous attributes for more accurate imputation.
4) For discrete attribute imputation (Type $\kappa$), \textsf{MH-GIN} reduces error rates by \warning{14.11\%--51.18\%} compared to Multi-task AIS. This superior performance stems from \textsf{MH-GIN}'s comprehensive modeling approach that incorporates all 12 attributes across different time scales, whereas Multi-task AIS only utilizes limited information (spatio-temporal attributes and vessel type). By effectively capturing cross-scale dependencies among the heterogeneous attributes, \textsf{MH-GIN} establishes more robust correlations that significantly enhance vessel type imputation accuracy.

\begin{table*}[!htbp]
    \small
    \centering
    \setlength{\tabcolsep}{0.9pt}
    \caption{{Effectiveness analysis of model components.}}
    \vspace{-1mm}
    \label{tab:ablation_components}
    \begin{tabular}{l|c|c|cc|cc|cc|cc|c|c|cc|cc|cc|c}
    \hline
\multicolumn{2}{c|}{\multirow{3}{*}{\makecell[c]{\textbf{Model}\\\textbf{Variant}}}} & \makecell[c]{Coord. $(\lambda,\phi)$} & \multicolumn{2}{c|}{\makecell[c]{Time $\tau$}} & \multicolumn{2}{c|}{\makecell[c]{Head. $\psi$}} & \multicolumn{2}{c|}{\makecell[c]{Course $\theta$}} & \multicolumn{2}{c|}{\makecell[c]{Speed $s$}} & \makecell[c]{Nav. $\eta$} & \makecell[c]{Cargo $\chi$} & \multicolumn{2}{c|}{\makecell[c]{Draft $d$}} & \multicolumn{2}{c|}{\makecell[c]{Length $\ell$}} & \multicolumn{2}{c|}{\makecell[c]{Width $\beta$}} & \makecell[c]{Type $\kappa$} \\
\cmidrule{3-20}
\multicolumn{2}{c|}{} & Dist. & MAE & SMAPE & MAE & SMAPE & MAE & SMAPE & MAE & SMAPE & ACC & ACC & MAE & SMAPE & MAE & SMAPE & MAE & SMAPE & ACC \\
    \midrule
    \multirow{9}{*}{\rotatebox{90}{AIS-DK}} 
    & \warninge{w/o Spa} & \warninge{1.364e-3} & \warninge{45.326} & \warninge{0.465} & \warninge{0.072} & \warninge{0.046} & \warninge{0.171} & \warninge{0.085} & \warninge{0.242} & \warninge{0.283} & \warninge{0.802} & \warninge{0.718} & \warninge{1.323} & \warninge{0.174} & \warninge{42.329} & \warninge{0.223} & \warninge{6.258} & \warninge{0.216} & \warninge{0.584} \\
    & \warninge{w/o Cyc} & \warninge{1.316e-4} & \warninge{25.769} & \warninge{0.158} & \warninge{0.062} & \warninge{0.041} & \warninge{0.176} & \warninge{0.269} & \warninge{0.216} & \warninge{0.261} & \warninge{0.881} & \warninge{0.769} & \warninge{1.134} & \warninge{0.143} & \warninge{35.106} & \warninge{0.188} & \warninge{4.994} & \warninge{0.184} & \warninge{0.651} \\
    & \warninge{w/o Con} & \warninge{1.151e-4} & \warninge{22.312} & \warninge{0.138} & \warninge{0.059} & \warninge{0.041} & \warninge{0.147} & \warninge{0.079} & \warninge{0.211} & \warninge{0.261} & \warninge{0.882} & \warninge{0.771} & \warninge{1.126} & \warninge{0.139} & \warninge{35.364} & \warninge{0.216} & \warninge{5.124} & \warninge{0.196} & \warninge{0.659} \\
    & \warninge{w/o Dis} & \warninge{1.249e-4} & \warninge{22.631} & \warninge{0.141} & \warninge{0.059} & \warninge{0.042} & \warninge{0.156} & \warninge{0.088} & \warninge{0.238} & \warninge{0.269} & - & - & \warninge{1.159} & \warninge{0.152} & \warninge{35.399} & \warninge{0.189} & \warninge{5.043} & \warninge{0.199} & - \\
    & \warninge{w/o M2-B} & \warninge{1.446e-4} & \warninge{28.643} & \warninge{0.179} & \warninge{0.069} & \warninge{0.053} & \warninge{0.192} & \warninge{0.106} & \warninge{0.246} & \warninge{0.286} & \warninge{0.813} & \warninge{0.625} & \warninge{1.249} & \warninge{0.171} & \warninge{38.352} & \warninge{0.204} & \warninge{5.984} & \warninge{0.215} & \warninge{0.532} \\
    & \warninge{w/o M3-S1} & \warninge{1.530e-4} & \warninge{31.825} & \warninge{0.372} & \warninge{0.872} & \warninge{0.456} & \warninge{0.185} & \warninge{0.102} & \warninge{0.758} & \warninge{0.410} & \warninge{0.322} & \warninge{0.380} & \warninge{1.385} & \warninge{0.765} & - & - & - & - & - \\
    & \warninge{w/o M3-S1-D} & \warninge{1.117e-4} & \warninge{21.652} & \warninge{0.129} & \warninge{0.050} & \warninge{0.037} & \warninge{0.143} & \warninge{0.074} & \warninge{0.203} & \warninge{0.256} & \warninge{0.898} & \warninge{0.792} & \warninge{1.019} & \warninge{0.127} & \warninge{34.388} & \warninge{0.180} & \warninge{4.901} & \warninge{0.178} & \warninge{0.665} \\
    & \warninge{w/o M3-S2} & \warninge{1.256e-4} & \warninge{23.125} & \warninge{0.142} & \warninge{0.051} & \warninge{0.038} & \warninge{0.145} & \warninge{0.076} & \warninge{0.206} & \warninge{0.257} & \warninge{0.890} & \warninge{0.780} & \warninge{1.095} & \warninge{0.135} & \warninge{34.825} & \warninge{0.183} & \warninge{4.950} & \warninge{0.180} & \warninge{0.658} \\
    & \warninge{MH-GIN}    & \warninge{\textbf{1.112e-4}} & \warninge{\textbf{21.413}} & \warninge{\textbf{0.127}} & \warninge{\textbf{0.049}} & \warninge{\textbf{0.036}} & \warninge{\textbf{0.141}} & \warninge{\textbf{0.071}} & \warninge{\textbf{0.198}} & \warninge{\textbf{0.252}} & \warninge{\textbf{0.904}} & \warninge{\textbf{0.798}} & \warninge{\textbf{1.017}} & \warninge{\textbf{0.126}} & \warninge{\textbf{34.192}} & \warninge{\textbf{0.178}} & \warninge{\textbf{4.896}} & \warninge{\textbf{0.175}} & \warninge{\textbf{0.669}} \\
    \midrule
    \multirow{9}{*}{\rotatebox{90}{AIS-US}} 
    & \warninge{w/o Spa} & \warninge{1.465e-4} & \warninge{28.59} & \warninge{0.082} & \warninge{0.132} & \warninge{0.125} & \warninge{0.493} & \warninge{0.239} & \warninge{0.469} & \warninge{1.326} & \warninge{0.592} & \warninge{0.524} & \warninge{2.258} & \warninge{0.349} & \warninge{41.536} & \warninge{0.329} & \warninge{5.869} & \warninge{0.373} & \warninge{0.394} \\
    & \warninge{w/o Cyc} & \warninge{9.242e-5} & \warninge{17.511} & \warninge{0.038} & \warninge{0.127} & \warninge{0.093} & \warninge{0.513} & \warninge{0.243} & \warninge{0.379} & \warninge{1.196} & \warninge{0.631} & \warninge{0.542} & \warninge{1.931} & \warninge{0.324} & \warninge{38.996} & \warninge{0.281} & \warninge{5.412} & \warninge{0.219} & \warninge{0.426} \\
    & \warninge{w/o Con} & \warninge{8.624e-5} & \warninge{15.866} & \warninge{0.018} & \warninge{0.119} & \warninge{0.085} & \warninge{0.479} & \warninge{0.189} & \warninge{0.382} & \warninge{1.191} & \warninge{0.633} & \warninge{0.545} & \warninge{1.965} & \warninge{0.347} & \warninge{39.26} & \warninge{0.284} & \warninge{5.762} & \warninge{0.373} & \warninge{0.419} \\
    & \warninge{w/o Dis} & \warninge{8.963e-5} & \warninge{15.976} & \warninge{0.021} & \warninge{0.127} & \warninge{0.082} & \warninge{0.483} & \warninge{0.189} & \warninge{0.386} & \warninge{1.196} & - & - & \warninge{1.953} & \warninge{0.367} & \warninge{39.104} & \warninge{0.292} & \warninge{5.536} & \warninge{0.227} & - \\
    & \warninge{w/o M2-B} & \warninge{1.265e-4} & \warninge{17.937} & \warninge{0.025} & \warninge{0.134} & \warninge{0.092} & \warninge{0.579} & \warninge{0.231} & \warninge{0.439} & \warninge{1.368} & \warninge{0.582} & \warninge{0.446} & \warninge{2.218} & \warninge{0.374} & \warninge{41.726} & \warninge{0.314} & \warninge{5.936} & \warninge{0.252} & \warninge{0.364} \\
    & \warninge{w/o M3-S1} & \warninge{1.372e-4} & \warninge{26.325} & \warninge{0.061} & \warninge{1.142} & \warninge{0.412} & \warninge{1.590} & \warninge{0.538} & \warninge{1.252} & \warninge{1.178} & \warninge{0.175} & \warninge{0.128} & \warninge{2.240} & \warninge{0.375} & - & - & - & - & - \\
    & \warninge{w/o M3-S1-D} & \warninge{8.364e-5} & \warninge{15.259} & \warninge{0.017} & \warninge{0.108} & \warninge{0.075} & \warninge{0.467} & \warninge{0.186} & \warninge{0.370} & \warninge{1.184} & \warninge{0.644} & \warninge{0.555} & \warninge{1.910} & \warninge{0.319} & \warninge{38.681} & \warninge{0.279} & \warninge{5.388} & \warninge{0.215} & \warninge{0.433} \\
    & \warninge{w/o M3-S2} & \warninge{8.731e-5} & \warninge{15.553} & \warninge{0.018} & \warninge{0.109} & \warninge{0.076} & \warninge{0.468} & \warninge{0.187} & \warninge{0.372} & \warninge{1.186} & \warninge{0.641} & \warninge{0.552} & \warninge{1.918} & \warninge{0.320} & \warninge{38.811} & \warninge{0.231} & \warninge{5.391} & \warninge{0.224} & \warninge{0.431} \\
    & \warninge{\textsf{MH-GIN}} & \warninge{\textbf{8.281e-5}} & \warninge{\textbf{15.146}} & \warninge{\textbf{0.016}} & \warninge{\textbf{0.104}} & \warninge{\textbf{0.072}} & \warninge{\textbf{0.461}} & \warninge{\textbf{0.184}} & \warninge{\textbf{0.368}} & \warninge{\textbf{1.182}} & \warninge{\textbf{0.651}} & \warninge{\textbf{0.559}} & \warninge{\textbf{1.899}} & \warninge{\textbf{0.314}} & \warninge{\textbf{38.263}} & \warninge{\textbf{0.273}} & \warninge{\textbf{5.382}} & \warninge{\textbf{0.213}} & \warninge{\textbf{0.446}} \\
    \hline
    \end{tabular}
\end{table*}
\begin{table*}
    \small
    \setlength{\tabcolsep}{1.6pt}
    \caption{{Robustness evaluation under different missing ratios.}}
    \vspace{-2mm}
    \label{tab:ablation_missingrate}
    \begin{tabular}{c|c|c|cc|cc|cc|cc|c|c|cc|cc|cc|c}
    \toprule
    \multicolumn{2}{c|}{\multirow{3}{*}{\makecell[c]{\textbf{Missing}\\\textbf{Rate}}}} & \makecell[c]{Coord. $(\lambda,\phi)$} & \multicolumn{2}{c|}{\makecell[c]{Time $\tau$}} & \multicolumn{2}{c|}{\makecell[c]{Head. $\psi$}} & \multicolumn{2}{c|}{\makecell[c]{Course $\theta$}} & \multicolumn{2}{c|}{\makecell[c]{Speed $s$}} & \makecell[c]{Nav. $\eta$} & \makecell[c]{Cargo $\chi$} & \multicolumn{2}{c|}{\makecell[c]{Draft $d$}} & \multicolumn{2}{c|}{\makecell[c]{Length $\ell$}} & \multicolumn{2}{c|}{\makecell[c]{Width $\beta$}} & \makecell[c]{Type $\kappa$} \\
    \cmidrule{3-20}
    \multicolumn{2}{c|}{} & Dist. & MAE & SMAPE & MAE & SMAPE & MAE & SMAPE & MAE & SMAPE & ACC & ACC & MAE & SMAPE & MAE & SMAPE & MAE & SMAPE & ACC \\
    \cmidrule{1-20}
    \multirow{5}{*}{\rotatebox{90}{AIS-DK}} 
    & 10\% & \warninge{0.979e-5} & \warninge{19.319} & \warninge{0.114} & \warninge{0.045} & \warninge{0.033} & \warninge{0.136} & \warninge{0.070} & \warninge{0.197} & \warninge{0.247} & \warninge{0.917} & \warninge{0.814} & \warninge{0.883} & \warninge{0.121} & \warninge{32.951} & \warninge{0.173} & \warninge{4.536} & \warninge{0.165} & \warninge{0.689} \\
    & 20\% & \warninge{1.043e-4} & \warninge{20.475} & \warninge{0.121} & \warninge{0.046} & \warninge{0.034} & \warninge{0.139} & \warninge{0.071} & \warninge{0.198} & \warninge{0.249} & \warninge{0.908} & \warninge{0.805} & \warninge{0.901} & \warninge{0.124} & \warninge{33.674} & \warninge{0.176} & \warninge{4.717} & \warninge{0.170} & \warninge{0.682} \\
    & 30\% & \warninge{1.112e-4} & \warninge{21.413} & \warninge{0.127} & \warninge{0.049} & \warninge{0.036} & \warninge{0.141} & \warninge{0.071} & \warninge{0.198} & \warninge{0.252} & \warninge{0.904} & \warninge{0.798} & \warninge{1.017} & \warninge{0.126} & \warninge{34.192} & \warninge{0.178} & \warninge{4.896} & \warninge{0.175} & \warninge{0.669} \\
    & 40\% & \warninge{3.229e-4} & \warninge{23.121} & \warninge{0.133} & \warninge{0.051} & \warninge{0.037} & \warninge{0.146} & \warninge{0.075} & \warninge{0.206} & \warninge{0.261} & \warninge{0.887} & \warninge{0.717} & \warninge{1.196} & \warninge{0.141} & \warninge{35.462} & \warninge{0.184} & \warninge{5.096} & \warninge{0.183} & \warninge{0.652} \\
    & 50\% & \warninge{1.383e-3} & \warninge{24.868} & \warninge{0.147} & \warninge{0.055} & \warninge{0.041} & \warninge{0.151} & \warninge{0.078} & \warninge{0.213} & \warninge{0.267} & \warninge{0.868} & \warninge{0.639} & \warninge{1.453} & \warninge{0.152} & \warninge{36.842} & \warninge{0.191} & \warninge{5.319} & \warninge{0.191} & \warninge{0.628} \\
    \midrule
    \multirow{5}{*}{\rotatebox{90}{AIS-US}} 
    & 10\% & \warninge{7.119e-5} & \warninge{14.313} & \warninge{0.014} & \warninge{0.101} & \warninge{0.069} & \warninge{0.451} & \warninge{0.177} & \warninge{0.356} & \warninge{1.164} & \warninge{0.664} & \warninge{0.573} & \warninge{1.838} & \warninge{0.306} & \warninge{37.536} & \warninge{0.271} & \warninge{5.239} & \warninge{0.206} & \warninge{0.456} \\
    & 20\% & \warninge{7.776e-5} & \warninge{14.822} & \warninge{0.015} & \warninge{0.103} & \warninge{0.072} & \warninge{0.456} & \warninge{0.181} & \warninge{0.362} & \warninge{1.174} & \warninge{0.659} & \warninge{0.563} & \warninge{1.875} & \warninge{0.313} & \warninge{38.109} & \warninge{0.273} & \warninge{5.304} & \warninge{0.210} & \warninge{0.449} \\
    & 30\% & \warninge{8.281e-5} & \warninge{15.146} & \warninge{0.016} & \warninge{0.104} & \warninge{0.072} & \warninge{0.461} & \warninge{0.184} & \warninge{0.368} & \warninge{1.182} & \warninge{0.651} & \warninge{0.559} & \warninge{1.899} & \warninge{0.314} & \warninge{38.263} & \warninge{0.273} & \warninge{5.382} & \warninge{0.213} & \warninge{0.446} \\
    & 40\% & \warninge{4.113e-4} & \warninge{15.861} & \warninge{0.017} & \warninge{0.110} & \warninge{0.077} & \warninge{0.473} & \warninge{0.189} & \warninge{0.372} & \warninge{1.192} & \warninge{0.634} & \warninge{0.546} & \warninge{1.962} & \warninge{0.324} & \warninge{39.246} & \warninge{0.283} & \warninge{5.469} & \warninge{0.219} & \warninge{0.424} \\
    & 50\% & \warninge{9.019e-4} & \warninge{16.533} & \warninge{0.018} & \warninge{0.115} & \warninge{0.081} & \warninge{0.484} & \warninge{0.194} & \warninge{0.385} & \warninge{1.206} & \warninge{0.621} & \warninge{0.529} & \warninge{2.023} & \warninge{0.333} & \warninge{40.115} & \warninge{0.291} & \warninge{5.579} & \warninge{0.225} & \warninge{0.411} \\
    \bottomrule
    \end{tabular}
\end{table*}
\subsubsection{Efficiency Analysis}
Table \ref{tab:efficiency_comparison} presents a comparison of computational efficiency in terms of inference time on the test dataset and memory consumption (model size) for each method. Key observations include:
1) Statistical methods (MEAN, KNN, Lin-ITP) achieve significantly lower computational times than parametric approaches.
2) \textsf{MH-GIN} demonstrates superior efficiency among neural methods, with inference times of 10.9s and 11.2s (22.7\% and 27.4\% faster than ImputeFormer) on AIS-DK and AIS-US datasets.
3) \textsf{MH-GIN} maintains minimal memory costs at 1.47MB and 1.92 MB. These efficiency gains come from  the Multi-scale Dependency Mining Module's efficient and simple architecture that eliminates complex operations required by complex neural networks.
\vspace{-2mm}
\subsection{Ablation Study}
To evaluate the contribution of each component of \textsf{MH-GIN}, we conducted a comprehensive ablation study by removing or replacing key components and analyzing the impact of different attribute types. Table \ref{tab:ablation_components} presents the results on model components. 

\begin{table*}
\small
\setlength{\tabcolsep}{1.3pt}
\caption{\warninge{Robustness evaluation under noisy conditions.}}
\label{tab:robustness_results}
\begin{tabular}{c|c|c|cc|cc|cc|cc|c|c|cc|cc|cc|c}
\toprule
\multicolumn{2}{c|}{\multirow{3}{*}{\makecell[c]{\textbf{Noisy}\\\textbf{Intensity}}}} & \makecell[c]{Coord. $(\lambda,\phi)$} & \multicolumn{2}{c|}{\makecell[c]{Time $\tau$}} & \multicolumn{2}{c|}{\makecell[c]{Head. $\psi$}} & \multicolumn{2}{c|}{\makecell[c]{Course $\theta$}} & \multicolumn{2}{c|}{\makecell[c]{Speed $s$}} & \makecell[c]{Nav. $\eta$} & \makecell[c]{Cargo $\chi$} & \multicolumn{2}{c|}{\makecell[c]{Draft $d$}} & \multicolumn{2}{c|}{\makecell[c]{Length $\ell$}} & \multicolumn{2}{c|}{\makecell[c]{Width $\beta$}} & \makecell[c]{Type $\kappa$} \\
\cmidrule{3-20}
\multicolumn{2}{c|}{} & Dist. & MAE & SMAPE & MAE & SMAPE & MAE & SMAPE & MAE & SMAPE & ACC & ACC & MAE & SMAPE & MAE & SMAPE & MAE & SMAPE & ACC \\
\cmidrule{1-20}
\multirow{5}{*}{\rotatebox{90}{AIS-DK}} 
& 0.00 & \warninge{1.112e-4} & \warninge{21.413} & \warninge{0.127} & \warninge{0.049} & \warninge{0.036} & \warninge{0.141} & \warninge{0.071} & \warninge{0.198} & \warninge{0.252} & \warninge{0.904} & \warninge{0.798} & \warninge{1.017} & \warninge{0.126} & \warninge{34.192} & \warninge{0.178} & \warninge{4.896} & \warninge{0.175} & \warninge{0.669} \\
& 0.01  & \warninge{1.123e-4} & \warninge{22.017} & \warninge{0.129} & \warninge{0.049} & \warninge{0.038} & \warninge{0.143} & \warninge{0.073} & \warninge{0.203} & \warninge{0.253} & \warninge{0.902} & \warninge{0.798} & \warninge{1.056} & \warninge{1.129} & \warninge{34.384} & \warninge{0.180} & \warninge{4.912} & \warninge{0.178} & \warninge{0.667}\\
& 0.015 & \warninge{1.366e-4} & \warninge{22.656} & \warninge{0.130} & \warninge{0.051} & \warninge{0.039} & \warninge{0.144} & \warninge{0.076} & \warninge{0.205} & \warninge{0.256} & \warninge{0.899} & \warninge{0.796} & \warninge{1.213} & \warninge{1.132} & \warninge{34.681} & \warninge{0.181} & \warninge{4.947} & \warninge{0.179} & \warninge{0.665}\\
& 0.02  & \warninge{1.647e-4} & \warninge{22.931} & \warninge{0.132} & \warninge{0.053} & \warninge{0.041} & \warninge{0.146} & \warninge{0.078} & \warninge{0.208} & \warninge{0.258} & \warninge{0.898} & \warninge{0.795} & \warninge{1.239} & \warninge{1.136} & \warninge{35.152} & \warninge{0.183} & \warninge{4.973} & \warninge{0.182} & \warninge{0.662}\\
& 0.025 & \warninge{2.148e-4} & \warninge{23.172} & \warninge{0.133} & \warninge{0.054} & \warninge{0.043} & \warninge{0.149} & \warninge{0.081} & \warninge{0.210} & \warninge{0.261} & \warninge{0.897} & \warninge{0.792} & \warninge{1.305} & \warninge{1.142} & \warninge{35.503} & \warninge{0.186} & \warninge{5.062} & \warninge{0.185} & \warninge{0.658}\\
\midrule
\multirow{5}{*}{\rotatebox{90}{AIS-US}} 
& 0.00 & \warninge{8.281e-5} & \warninge{15.146} & \warninge{0.016} & \warninge{0.104} & \warninge{0.072} & \warninge{0.461} & \warninge{0.184} & \warninge{0.368} & \warninge{1.182} & \warninge{0.651} & \warninge{0.559} & \warninge{1.899} & \warninge{0.314} & \warninge{38.263} & \warninge{0.273} & \warninge{5.382} & \warninge{0.213} & \warninge{0.446} \\
& 0.01 & \warninge{8.316e-5} & \warninge{15.265} & \warninge{0.016} & \warninge{0.106} & \warninge{0.074} & \warninge{0.463} & \warninge{0.188} & \warninge{0.369} & \warninge{1.184} & \warninge{0.648} & \warninge{0.557} & \warninge{1.912} & \warninge{0.316} & \warninge{38.426} & \warninge{0.275} & \warninge{5.399} & \warninge{0.214} & \warninge{0.445} \\
& 0.015 & \warninge{8.999e-5} & \warninge{15.382} & \warninge{0.016} & \warninge{0.107} & \warninge{0.076} & \warninge{0.466} & \warninge{0.190} & \warninge{0.372} & \warninge{1.188} & \warninge{0.645} & \warninge{0.554} & \warninge{1.946} & \warninge{0.319} & \warninge{38.873} & \warninge{0.278} & \warninge{5.402} & \warninge{0.217} & \warninge{0.443}\\
& 0.02 & \warninge{9.642e-5} & \warninge{15.503} & \warninge{0.017} & \warninge{0.109} & \warninge{0.079} & \warninge{0.468} & \warninge{0.192} & \warninge{0.375} & \warninge{1.191} & \warninge{0.643} & \warninge{0.552} & \warninge{1.980} & \warninge{0.321} & \warninge{39.143} & \warninge{0.281} & \warninge{5.436} & \warninge{0.219} & \warninge{0.441}\\
& 0.025 & \warninge{1.017e-4} & \warninge{15.621} & \warninge{0.017} & \warninge{0.112} & \warninge{0.081} & \warninge{0.472} & \warninge{0.195} & \warninge{0.379} & \warninge{1.195} & \warninge{0.640} & \warninge{0.549} & \warninge{2.014} & \warninge{0.325} & \warninge{39.479} & \warninge{0.284} & \warninge{5.475} & \warninge{0.222} & \warninge{0.438}\\
\bottomrule
\end{tabular}
\end{table*}
\begin{table*}
    \small
    \setlength{\tabcolsep}{1.6pt}
    \caption{\warninge{Effectiveness of MH-GIN in scenario where multiple attributes are missing at the same time.}}
    \vspace{-2mm}
    \label{tab:multi_missing_value}
    \begin{tabular}{c|c|c|cc|cc|cc|cc|c|c|cc|cc|cc|c}
    \toprule
    \multicolumn{2}{c|}{\multirow{3}{*}{\makecell[c]{\textbf{Missing} \\ \textbf{Number}}}} & \makecell[c]{Coord. $(\lambda,\phi)$} & \multicolumn{2}{c|}{\makecell[c]{Time $\tau$}} & \multicolumn{2}{c|}{\makecell[c]{Head. $\psi$}} & \multicolumn{2}{c|}{\makecell[c]{Course $\theta$}} & \multicolumn{2}{c|}{\makecell[c]{Speed $s$}} & \makecell[c]{Nav. $\eta$} & \makecell[c]{Cargo $\chi$} & \multicolumn{2}{c|}{\makecell[c]{Draft $d$}} & \multicolumn{2}{c|}{\makecell[c]{Length $\ell$}} & \multicolumn{2}{c|}{\makecell[c]{Width $\beta$}} & \makecell[c]{Type $\kappa$} \\
    \cmidrule{3-20}
    \multicolumn{2}{c|}{} & Dist. & MAE & SMAPE & MAE & SMAPE & MAE & SMAPE & MAE & SMAPE & ACC & ACC & MAE & SMAPE & MAE & SMAPE & MAE & SMAPE & ACC \\
    \cmidrule{1-20}
    \multirow{3}{*}{\rotatebox{90}{AIS-DK}} 
    & 1-2 & \warninge{1.038e-4} & \warninge{21.316} & \warninge{0.113} & \warninge{0.045} & \warninge{0.031} & \warninge{0.132} & \warninge{0.061} & \warninge{0.193} & \warninge{0.241} & \warninge{0.916} & \warninge{0.807} & \warninge{0.987} & \warninge{0.119} & \warninge{33.992} & \warninge{0.162} & \warninge{4.632} & \warninge{0.168} & \warninge{0.681} \\
    & 3-4 & \warninge{1.108e-4} & \warninge{21.388} & \warninge{0.125} & \warninge{0.047} & \warninge{0.034} & \warninge{0.139} & \warninge{0.067} & \warninge{0.196} & \warninge{0.246} & \warninge{0.911} & \warninge{0.802} & \warninge{1.004} & \warninge{0.123} & \warninge{34.014} & \warninge{0.167} & \warninge{4.784} & \warninge{0.171} & \warninge{0.673} \\
    & 5-6 & \warninge{1.726e-4} & \warninge{22.632} & \warninge{0.139} & \warninge{0.056} & \warninge{0.044} & \warninge{0.146} & \warninge{0.078} & \warninge{0.207} & \warninge{0.259} & \warninge{0.892} & \warninge{0.784} & \warninge{1.126} & \warninge{0.141} & \warninge{34.369} & \warninge{0.183} & \warninge{4.996} & \warninge{0.185} & \warninge{0.661} \\
    \midrule
    \multirow{3}{*}{\rotatebox{90}{AIS-US}} 
    & 1-2 & \warninge{8.103e-5} & \warninge{14.936} & \warninge{0.013} & \warninge{0.091} & \warninge{0.065} & \warninge{0.448} & \warninge{0.178} & \warninge{0.356} & \warninge{1.169} & \warninge{0.661} & \warninge{0.569} & \warninge{1.821} & \warninge{0.298} & \warninge{38.184} & \warninge{0.253} & \warninge{5.324} & \warninge{0.199} & \warninge{0.454} \\
    & 3-4 & \warninge{8.226e-5} & \warninge{15.012} & \warninge{0.015} & \warninge{0.096} & \warninge{0.070} & \warninge{0.452} & \warninge{0.181} & \warninge{0.359} & \warninge{1.173} & \warninge{0.657} & \warninge{0.562} & \warninge{1.832} & \warninge{0.302} & \warninge{38.226} & \warninge{0.261} & \warninge{5.371} & \warninge{0.208} & \warninge{0.449} \\
    & 5-6 & \warninge{8.394e-5} & \warninge{15.233} & \warninge{0.019} & \warninge{0.116} & \warninge{0.078} & \warninge{0.477} & \warninge{0.189} & \warninge{0.371} & \warninge{1.189} & \warninge{0.648} & \warninge{0.554} & \warninge{1.904} & \warninge{0.323} & \warninge{38.284} & \warninge{0.276} & \warninge{5.394} & \warninge{0.226} & \warninge{0.441} \\
    \bottomrule
    \end{tabular}
\end{table*}
We have following observations:
1) w/o Spa, w/o Cyc, w/o Con, w/o Dis: Removing type-specific encoders and decoders for spatio-temporal, cyclical, continuous, and discrete attributes leads to significant performance degradation in their corresponding attribute types. Performance also drops for other attribute types due to missing well-encoded representations, demonstrating the interdependencies between heterogeneous attributes. 
2) w/o M2-B: Replacing bidirectional DeepESN with forward-only variant reduces performance across all attributes. Nevertheless, M2-B still outperforms the baseline methods (as shown in Table~\ref{tab:performance_comparison-AIS-DK} and Table~\ref{tab:performance_comparison-AIS-US}), demonstrating the effectiveness of \textsf{MH-GIN} in real-time scenarios. 
3) w/o M3-S1: Removing intra-scale propagation severely impacts performance across all attributes, demonstrating that \textsf{MH-GIN} cannot effectively capture inter-attribute dependencies without this component. This highlights the critical importance of modeling complementary relationships between heterogeneous attributes within each time scale.
4) w/o M3-S1-D: Removing dynamic edges within time-scale subgraphs while keeping static graph structure shows performance degradation compared to the full model, demonstrating that dynamic relationship modeling is essential for capturing the evolving dependencies between attributes at each time scale.
5) w/o M3-S2: Eliminating cross-scale propagation results in performance degradation, particularly for attributes with dependencies across scales. This demonstrates that cross-scale relationships are essential for effectively leveraging complementary information across different temporal dynamics. 


\subsection{Robustness of MH-GIN}
\noindent \textbf{Performance under Different Missing Ratios.}
We analyze \textsf{MH-GIN}'s robustness under varying missing ratios (10\%-50\%) in Table~\ref{tab:ablation_missingrate}. The results yield the following observations:
1) Performance degrades gradually with increasing missing ratios across all attributes, indicating model stability under data sparsity.
2) Non-coordinate attributes exhibit near-linear degradation without catastrophic collapse. The hierarchical structure enables lower-scale attributes to generate higher-scale features, allowing the model to leverage abundant high-scale features for reasonable imputation performance.
3) Coordinate attributes (lowest time scale) show significant performance drops at high missing ratios due to insufficient low-scale features and limited constraints from coarse high-scale features. However, such extreme sparsity is rare in practice, where coordinates typically remain densely observed while non-spatio-temporal attributes exhibit higher missing ratios.
\textsf{MH-GIN} maintains strong performance under high missing ratios, demonstrating robustness in practical settings.

\noindent \textbf{Performance under Different Noise Intensities.}
To evaluate the robustness of \textsf{MH-GIN} under noisy conditions, we inject noise into the AIS data with progressively varying noise intensity $\gamma$ to simulate realistic data corruption scenarios. As demonstrated in Table~\ref{tab:robustness_results}, \textsf{MH-GIN} exhibits remarkable resilience across noise intensities ranging from 0.0 to 0.025. The model exhibits graceful performance degradation in a near-linear fashion without catastrophic failure, which can be attributed to the multi-scale architecture's capability to leverage contextual information from less corrupted attributes and time scales, thereby effectively mitigating the impact of local perturbations.

\noindent \textbf{Performance under Multiple Attributes Missing Scenarios.}
To evaluate \textsf{MH-GIN}'s robustness under simultaneous multi-attribute missing scenarios, we assess the model's performance as the number of missing attributes increases. As shown in \warning{Table~\ref{tab:multi_missing_value}}, while MH-GIN's performance degrades as the number of missing attributes increases, the degradation remains moderate and controlled.

\section{Conclusion and Future Work}
\label{sec:conclusion}
Our study highlights the importance of modeling multi-scale dependencies between heterogeneous attributes for missing attribute imputation in AIS data. 
We introduce \textsf{MH-GIN}, a Multi-scale Heterogeneous Graph-based Imputation Network for AIS data. \textsf{MH-GIN} is the first method that can effectively model multi-scale dependencies between heterogeneous attributes for missing attribute imputation in AIS data. 
\textsf{MH-GIN} first captures the multi-scale temporal features of each attribute while preserving their intrinsic properties. It then constructs a multi-scale heterogeneous graph to model the complex dependencies between attributes, enabling effective missing attribute imputation through two-stage graph propagation.
Experiments on two real-world AIS datasets demonstrate \textsf{MH-GIN}'s superior performance over state-of-the-art methods. 
Experiments on two real-world AIS datasets demonstrate \textsf{MH-GIN}'s superior performance over state-of-the-art methods across all attribute types, achieving an average 57\% reduction in imputation error, while maintaining computational efficiency. 

In future work, we plan to extend \textsf{MH-GIN} to other domains with heterogeneous attributes across different time scales, such as financial data, medical data, and IoT sensor data.

\section{Acknowledgment}
The research leading to the results presented in this paper has received funding from the European Union’s funded Project MobiSpaces under grant agreement no 101070279.


\balance
\bibliographystyle{ACM-Reference-Format}

\bibliography{MH-GIN}

\clearpage
\appendix
\section{The Details of MH-GIN}
\subsection{Heterogeneous Attribute Encoder}
\label{sec:appendix_heterogeneous_attribute_encoder}
\noindent\textbf{Spatio-temporal Attribute Encoder.}
Spherical coordinates (longitude $\lambda$, latitude $\phi$, as defined in Definition~\ref{def:ais_record} and Table~\ref{tab:attribute_characteristics}) exhibit non-Euclidean geometry, as standard numeric encodings fails to capture spherical continuity and latitude-dependent distortion \cite{DBLP:conf/iclr/CohenGKW18}. We address this by extending 3D Cartesian projections \cite{wiki_cartesian_2023} with additional low-order harmonic terms:
\begin{equation}
\begin{aligned}
\mathbf{f}(\lambda, \phi) = [
 & \sin(\lambda)\cos(\phi),\;
   \cos(\lambda)\cos(\phi),\;
   \sin(\phi), \\
 & \sin(2\lambda)\cos(\phi),\;
   \cos(2\lambda)\cos(\phi)
]^T,
\end{aligned}
\end{equation}
where \(\lambda\) and \(\phi\) are in radians. The first three components correspond to the standard 3D Cartesian mapping, while the two additional terms enhance directional sensitivity. We then apply a learnable affine transform and a hyperbolic tangent:
\begin{equation}
\begin{aligned}
\mathbf{e}_{\lambda} &= \tanh\!\bigl(\mathbf{W}_{\lambda} \cdot \mathbf{f}(\lambda,\phi) + b_{\lambda}\bigr), \\
\mathbf{e}_{\phi} &= \tanh\!\bigl(\mathbf{W}_{\phi} \cdot \mathbf{f}(\lambda,\phi) + b_{\phi}\bigr),
\end{aligned}
\end{equation}
where \(\mathbf{W}_{\lambda}, \mathbf{W}_{\phi} \in \mathbb{R}^{d \times 5}\) and \(b_{\lambda}, b_{\phi} \in \mathbb{R}^d\) are learnable parameters.

Timestamp $\tau$ (as defined in Definition~\ref{def:ais_record} and Table~\ref{tab:attribute_characteristics}) has nested periodicities. To preserve it for neural processing, we employ multi-frequency sinusoidal encoding:
\begin{equation}
e_{\tau, i} = \begin{cases} 
\sin(2\pi \tau/\mathbf{T}_{\lfloor i/2 \rfloor+1}) & i \bmod 2 = 0, \\
\cos(2\pi \tau/\mathbf{T}_{\lfloor i/2 \rfloor+1}) & i \bmod 2 \neq 0,
\end{cases}
\end{equation}
where \(\mathbf{T} = \{24a, 168a, 720a, 8760a\}\) captures key cycles \cite{10.24963/ijcai.2024/803} (with \(a\) as the hour unit). These sinusoidal outputs then pass through a linear layer with the $\tanh$ activation function, producing a \(d\)-dimensional representation $\mathbf{e}_{\tau} \in\mathbb{R}^d$.

\noindent \textbf{Cyclical Attribute Encoder.}
Cyclical attributes $\mathbf{X}_c = \{\psi, \theta\}$ (as defined in Definition~\ref{def:ais_record}) exhibit boundary continuity requirements. To preserve rotational equivalence while enabling effective gradient flow, we implement trigonometric encoding with $\tanh$ activation function:
\begin{equation}
\label{eq:cyclical_encoder}
\mathbf{e}_c = \tanh\!\left(
\mathbf{W}_c \cdot 
\begin{bmatrix}
\sin(\pi x_c/180) \\
\cos(\pi x_c/180)
\end{bmatrix}
+ \mathbf{b}_c
\right),
\end{equation}
where the trigonometric basis explicitly encodes angular continuity \cite{schmidt2012modeling}, and the $\tanh$ activation function provides a non-linear transformation while maintaining boundary-sensitive gradient propagation. 
This formulation leverages trigonometric basis functions to maintain angular continuity in the embedding space $\mathbf{e}_c \in\mathbb{R}^d$, where neighboring angular values exhibit smooth transitions while preserving gradient stability.

\noindent\textbf{Continuous Attribute Encoder.}
Continuous attributes $\mathbf{X}_n = \{s, d, \ell, \beta\}$ (as defined in Definition~\ref{def:ais_record}) range widely in scale, which can skew training dynamics. We address this via adaptive normalization:
\begin{equation}
\hat{x}_n = \frac{x_n - \mu}{\sigma} \alpha + \beta,
\end{equation}
where $\mu,\sigma$ are running mean and standard deviation of attribute $x_n$ in training set, $x_n$ represents any attribute within the continuous attribute set $\mathbf{X}_n$, $\alpha,\beta$ are learnable parameters of attribute $x_n$. Finally, a linear layer with the ReLU activation function yields:
\begin{equation}
\mathbf{e}_n = \mathrm{ReLU}(\mathbf{W}_n \cdot \hat{x}_n + \mathbf{b}_n),
\end{equation}
helping preserve original distributions while stabilizing gradients.

\noindent\textbf{Discrete Attribute Encoder.}
Discrete attributes $\mathbf{X}_d = \{\eta, \chi, \kappa\}$ (as defined in Definition~\ref{def:ais_record}) represent categorical semantics without intrinsic ordering. We first convert them to one-hot vectors, then project:
\begin{equation}
\mathbf{e}_d 
= \mathrm{tanh}(\mathbf{W}_d \cdot\mathrm{one\text{-}hot}(x_d)
+ \mathbf{b}_d),
\end{equation}
where \(\mathbf{W}_d \in \mathbb{R}^{d \times |C|}\), $|C|$ is the number of categories in attribute $x_d$, and $x_d$ represents any attribute within the discrete attribute set $\mathbf{X}_d$. Finally, all type-specific representations are concatenated into:
\[
\mathbf{E}_t 
= [\mathbf{E}_{s,t}; \,\mathbf{E}_{c,t}; \,\mathbf{E}_{n,t}; \,\mathbf{E}_{d,t}]
\in \mathbb{R}^{N \times d},
\]
preserving each attribute's intrinsic properties within a shared embedding space.

\subsection{Heterogeneous Attribute Imputation}
\label{sec:appendix_heterogeneous_attribute_imputation}
\noindent \textbf{Spatio-temporal Decoder.} 
For coordinate attributes, we leverage the inherent spatial continuity of vessel trajectories by using local averages as base estimates. This enables the network to focus on learning incremental adjustments rather than absolute positions.

Given missing coordinate attributes $\lambda_i$ and $\phi_i$ at time step $i$, we first compute the mean coordinates from nearby valid observations within a window:
\begin{equation}
\begin{aligned}
\mathcal{N}_i = \{(\lambda_j, \phi_j) \mid & |i - j| \leq \Delta, m_{\lambda_i}=1, m_{\phi_i}=1\},\\[1mm]
(\lambda_{\text{base}, i}, \phi_{\text{base}, i}) &= \frac{1}{|\mathcal{N}_i|}\sum_{(\lambda_j, \phi_j) \in \mathcal{N}_i} (\lambda_j, \phi_j),
\end{aligned}
\end{equation}
where $m_{\lambda_i}$ and $m_{\phi_i}$ represent the observation masks defined in Section~\ref{sec:problem_formulation}, and $\Delta$ specifies the temporal window size. The model then computes spatial adjustments:
\begin{equation}
\begin{aligned}
\delta_{\lambda} &= \tanh\big(\mathbf{W}_{\lambda} \cdot [\mathbf{\tilde{e}}_{\lambda_i}, \mathbf{\tilde{e}}_{\phi_i}] + \mathbf{b}_{\lambda}\big)  \gamma_{\lambda},\\[1mm]
\delta_{\phi} &= \tanh\big(\mathbf{W}_{\phi} \cdot [\mathbf{\tilde{e}}_{\lambda_i}, \mathbf{\tilde{e}}_{\phi_i}] + \mathbf{b}_{\phi}\big)  \gamma_{\phi},
\end{aligned}
\end{equation}
where $\gamma_{\lambda}$ and $\gamma_{\phi}$ are learnable scaling factors, and $\mathbf{\tilde{e}}_{\lambda_i}, \mathbf{\tilde{e}}_{\phi_i}$ are feature representations from Eq.~\ref{eq:gated_fusion}. The final imputed coordinates are:
\begin{equation}
\begin{aligned}
\hat{\lambda} = \lambda_{\text{base}} + \delta_{\lambda}, \quad \hat{\phi} = \phi_{\text{base}} + \delta_{\phi}
\end{aligned}
\end{equation}

For timestamp attribute \(\tau\), we predict time intervals between consecutive records rather than absolute timestamps, modeling temporal occurrences as a temporal point process \cite{lin2022exploring,omi2019fully}. 
We implement a conditional intensity function \(\eta(\tau)\) that combines a learnable base intensity \(\eta_0 > 0\) with a neural network function \(f(\mathbf{\tilde{e}}_{\tau})\), ensuring non-negativity through Softplus activation function:  
\begin{equation}
f_{\theta}(\mathbf{\tilde{e}}_{\tau}) = \text{Softplus}(\mathbf{W}_{\tau, 2} \cdot \text{SiLU}(\mathbf{W}_{\tau, 1} \cdot \mathbf{\tilde{e}}_{\tau} + \mathbf{b}_{\tau, 1}) + \mathbf{b}_{\tau, 2}),
\end{equation}  
where $\mathbf{W}_{\tau, 1}, \mathbf{W}_{\tau, 2} \in \mathbb{R}^{d \times d}$ are learnable weights, $\mathbf{b}_{\tau, 1}, \mathbf{b}_{\tau, 2} \in \mathbb{R}^d$ are bias vectors, and $\text{SiLU}(\cdot)$ is the self-gated activation function \cite{ramachandran2017searching}. The predicted timestamp is derived as:  
\begin{equation}
\hat{\tau}_{i} = \tau_{i-1} - \frac{\log(u)}{\eta(\tau)}, 
\end{equation}  
where $u \sim \text{Uniform}(0, 1)$.

\noindent \textbf{Cyclical Decoder.} 
To preserve cyclical continuity, the cyclical decoder reconstructs angular values using trigonometric projection. Specifically, it generates normalized trigonometric components:

\begin{equation}
\label{eq:cyclical_decoder}
\mathbf{h}_c = \mathbf{W}_{c,2} \cdot \tanh(\mathbf{W}_{c,1} \cdot \mathbf{\tilde{e}}_{x_c} + \mathbf{b}_{c,1}) + \mathbf{b}_{c,2}, \quad 
\hat{\mathbf{e}}_{c} = \frac{\mathbf{h}_c}{\|\mathbf{h}_c\|_2},
\end{equation}
where $\mathbf{W}_{c} \in \mathbb{R}^{2 \times d}$ represents learnable weights and $\mathbf{b}_{c} \in \mathbb{R}^2$ is a bias vector.
The $\ell_2$-normalization ensures $\hat{e}_{c,1}^2 + \hat{e}_{c,2}^2 = 1$, maintaining trigonometric identity. The angular value is then decoded as:
\begin{equation}
\hat{x}_c = \frac{180}{\pi} \arctan_2(\hat{e}_{c,1}, \hat{e}_{c,2}),
\end{equation}

\noindent \textbf{Continuous Decoder.} 
For continuous attributes, we transform the embedding into a normalized representation and then implement an inverse normalization transformation to restore original values:
\begin{equation}
\begin{aligned}
h_n &= \text{ReLU}(\mathbf{W}_{n} \cdot \mathbf{\tilde{e}}_{x_n} + \mathbf{b}_{n}), \\
\hat{x}_n &= \mu + \sigma \,\frac{h_n - \beta}{\alpha}, \\
\end{aligned}
\end{equation}
where $\mathbf{W}_{n} \in \mathbb{R}^{1 \times d}$ and $\mathbf{b}_{n} \in \mathbb{R}$ are learnable parameters, $\text{ReLU}(\cdot)$ is the Rectified Linear Unit activation function, $\mu,\sigma$ represent the attribute's statistical moments from training data, and $\alpha,\beta$ are the normalization parameters from the encoder (see Section~\ref{sec:encoding-module}).

\noindent\textbf{Discrete Decoder.} 
For discrete attributes, we apply a linear layer followed by a softmax to predict class probabilities:
\begin{equation}
\begin{aligned}
\hat{y} &= \text{softmax}(\mathbf{W}_d \cdot \mathbf{\tilde{e}}_{x_d} + \mathbf{b}_d), 
\end{aligned}
\end{equation}
where $\mathbf{W}_d \in \mathbb{R}^{|C| \times d}$ and $\mathbf{b}_d \in \mathbb{R}^{|C|}$ are learnable parameters, and $|C|$ is the number of discrete classes.

\subsection{Training Details}
\label{sec:appendix_training_details}
To optimize \textsf{MH-GIN}, we design a comprehensive loss function that combines specialized reconstruction losses for each heterogeneous attribute type:
\begin{equation}
\mathcal{L}_{\textit{total}} = \lambda_{\textit{coo}} \mathcal{L}_{\textit{coo}} + \lambda_{\tau} \mathcal{L}_{\tau} + \lambda_{\textit{period}} \mathcal{L}_{\textit{period}} + \lambda_{\textit{cont}} \mathcal{L}_{\textit{cont}} + \lambda_{\textit{disc}} \mathcal{L}_{\textit{disc}},
\end{equation}
where $\lambda_{\textit{coo}}$, $\lambda_{\tau}$, $\lambda_{\textit{period}}$, $\lambda_{\textit{cont}}$, and $\lambda_{\textit{disc}}$ are balancing hyperparameters with default values of 1. Each loss term corresponds to a specific attribute type: $\mathcal{L}_{\textit{coo}}$ for coordinates, $\mathcal{L}_{\tau}$ for timestamps, $\mathcal{L}_{\textit{period}}$ for cyclical attributes, $\mathcal{L}_{\textit{cont}}$ for continuous attributes, and $\mathcal{L}_{\textit{disc}}$ for discrete attributes.

\noindent\textbf{Spatio-temporal Loss.} 
For spatio-temporal attributes, we employ the Haversine distance \cite{wiki_haversine_2023} to measure the spherical distance between predicted coordinates and ground truth coordinates:
\begin{equation}
\mathcal{L}_{\textit{coo}} = 2  \arcsin\left(\sqrt{\sin^2\left(\frac{\hat{\phi} - \phi}{2}\right) + \cos(\phi)  \cos(\hat{\phi})  \sin^2\left(\frac{\hat{\lambda} - \lambda}{2}\right)}\right),
\end{equation}
where $(\hat{\lambda}, \hat{\phi})$ represents the predicted coordinates, $(\lambda, \phi)$ denotes the ground truth coordinates, and 
the distance is measured in radians to balance the magnitude of this loss term with other loss components.

For timestamp attributes, we utilize the mean squared error between the predicted time intervals and ground truth time intervals as the loss function:
\begin{equation}
\mathcal{L}_{\tau} = \frac{1}{N}\sum_{i=1}^{N}((\hat{\tau}_i - \tau_{i-1}) - (\tau_i - \tau_{i-1}))^2,
\end{equation}
where $\hat{\tau}_i$ represents the predicted timestamp, $\tau_i$ denotes the ground truth timestamp, and $(\tau_i - \tau_{i-1})$ and $(\hat{\tau}_i - \hat{\tau}_{i-1})$ are the actual and predicted time intervals between consecutive records.

\noindent\textbf{Cyclical Loss.} 
For cyclical attributes, instead of directly measuring the difference between attribute values, we compute the loss based on their trigonometric encodings (see  Eq.~\ref{eq:cyclical_encoder}) and the predicted trigonometric encodings (see Eq.~\ref{eq:cyclical_decoder}) to better capture the cyclical nature of these attributes:
\begin{equation}
\mathcal{L}_{\textit{cycl}} = \left\|\mathbf{e}_{c} - \mathbf{\hat{e}}_{c}\right\|_2,
\end{equation}
where $\mathbf{e}_{c}$ and $\mathbf{\hat{e}}_{c}$ denote the trigonometric encodings of ground-truth and predicted cyclical values. This loss enables the model to better learn the inherent cyclical patterns in cyclical attributes.

\noindent \textbf{Continuous Loss.} 
For continuous attributes, we apply mean squared error to measure reconstruction accuracy:
\begin{equation}
\mathcal{L}_{\mathrm{cont}}
\;=\;
\frac{1}{|\mathbf{x}_n|}
\sum_{x_n \in \mathbf{x}_n} 
(\hat{x}_n - x_n)^2,
\label{eq:continuous_loss}
\end{equation}
where $\hat{x}_n$ represents the predicted value and $x_n$ denotes the ground truth for each continuous attribute in $\mathbf{x}_n$.

\noindent\textbf{Discrete Loss.} For discrete attributes, we employ cross-entropy loss to quantify the divergence between predicted and true distributions:
\begin{equation}
\mathcal{L}_{\text{disc}} = -\sum_{i=1}^{|C|} y_i \log(\hat{y}_i),
\end{equation}
where $\hat{y}_i$ denotes the predicted probability for class $i$, and $y_i$ represents the $i$-th element of the one-hot encoded ground truth.

\section{PROOF OF LEMMAS}
\subsection{Stability of Multi-scale Propagation}
\label{sec:appendix_stability_of_multi_scale_propagation}
\begin{lemma}[Stability of Multi-scale Propagation]
The multi-scale propagation mechanism is numerically stable if the spectral radius of each propagation matrix satisfies $\rho(\mathbf{P}) = \mathbf{D}^{-1/2} \hat{\mathbf{A}} \mathbf{D}^{-1/2} \leq 1$.
\end{lemma}

\begin{proof}
We prove the stability by analyzing each stage of the propagation mechanism.

\noindent \textbf{Stage 1: Intra-scale Propagation Stability.}
For each time-scale subgraph $G^k$, the intra-scale propagation is defined as:
$$
\tilde{\mathbf{H}}^k = (\mathbf{D}^k)^{-1/2} \hat{\mathbf{A}}^k (\mathbf{D}^k)^{-1/2} \mathbf{H}^k = \mathbf{P}^k \mathbf{H}^k
$$
where $\mathbf{P}^k = (\mathbf{D}^k)^{-1/2} \hat{\mathbf{A}}^k (\mathbf{D}^k)^{-1/2}$ is the normalized propagation matrix. The symmetric normalization ensures that $\mathbf{P}^k$ is a doubly stochastic matrix when $\hat{\mathbf{A}}^k$ represents valid edge weights. By the Perron-Frobenius theorem~\cite{wiki_perron_frobenius_2024}, the spectral radius of such matrices is bounded: $\rho(\mathbf{P}^k) \leq 1$. This makes the intra-scale propagation a non-expansive mapping:
$$
\|\tilde{\mathbf{H}}^k\|_2 = \|\mathbf{P}^k \mathbf{H}^k\|_2 \leq \|\mathbf{P}^k\|_2 \|\mathbf{H}^k\|_2 = \rho(\mathbf{P}^k) \|\mathbf{H}^k\|_2 \leq \|\mathbf{H}^k\|_2
$$

\noindent \textbf{Stage 2: Cross-scale Propagation Stability.}
The cross-scale propagation for each attribute subgraph $G_x$ is defined as:
$$
\hat{\mathbf{H}}_x = (\mathbf{D}_x)^{-1/2} \hat{\mathbf{A}}_x (\mathbf{D}_x)^{-1/2} \tilde{\mathbf{H}}_x = \mathbf{P}_x \tilde{\mathbf{H}}_x
$$

Similar to the intra-scale case, the normalized matrix $\mathbf{P}_x$ satisfies $\rho(\mathbf{P}_x) \leq 1$, making this stage also non-expansive:
$$
\|\hat{\mathbf{H}}_x\|_2 = \|\mathbf{P}_x \tilde{\mathbf{H}}_x\|_2 \leq \|\tilde{\mathbf{H}}_x\|_2
$$

\noindent \textbf{Overall Stability.}
The complete two-stage propagation can be viewed as the composition of two non-expansive mappings. For any input $\mathbf{H}^k$, the final output satisfies:
$$
\|\hat{\mathbf{H}}_x\|_2 \leq \|\tilde{\mathbf{H}}_x\|_2 \leq \|\mathbf{H}^k\|_2
$$

This ensures that feature representations remain bounded throughout the propagation process, preventing numerical instability such as gradient explosion or vanishing. The iterative application of this mechanism converges to a stable fixed point, guaranteeing the numerical stability of the entire multi-scale propagation.
\end{proof}

\subsection{Robustness of Multi-scale Propagation}
\label{sec:appendix_robustness_of_multi_scale_propagation}
We analyze the model's robustness to input perturbations (e.g., from missing data or noise) by proving the Lipschitz continuity of the propagation operator. Let the full propagation operator be denoted by $\mathcal{G}$.

\begin{definition}[Lipschitz Continuity]
A function $g: \mathbb{R}^n \to \mathbb{R}^m$ is $L$-Lipschitz continuous if there exists a constant $L \ge 0$ such that for any two inputs $\mathbf{h}_1, \mathbf{h}_2 \in \mathbb{R}^n$, the inequality $\|g(\mathbf{h}_1) - g(\mathbf{h}_2)\| \le L \|\mathbf{h}_1 - \mathbf{h}_2\|$ holds~\cite{wiki_lipschitz_2025}.
\end{definition}

\begin{lemma}[Robustness of Multi-scale Propagation]
The multi-scale propagation operator $\mathcal{G}$ is Lipschitz continuous.
\end{lemma}

\begin{proof}
The intra-scale and cross-scale propagation steps are linear operators. The Lipschitz constant of a linear operator is its spectral norm. As established in the stability proof, the spectral norms of the propagation matrices are bounded by 1 (i.e., $\|\mathbf{P}^k\|_2 = \rho(\mathbf{P}^k) \le 1$). Therefore, both propagation steps are \textbf{1-Lipschitz}. For any two input features $\mathbf{H}$ and a perturbed version $\mathbf{H}'$, the output difference is bounded:
\[
\|\mathcal{G}_{\text{intra}}(\mathbf{H}) - \mathcal{G}_{\text{intra}}(\mathbf{H}')\|_2 \le 1 \cdot \|\mathbf{H} - \mathbf{H}'\|_2
\]

The overall propagation is a composition of these 1-Lipschitz functions. The composition of an $L_1$-Lipschitz function and an $L_2$-Lipschitz function is $(L_1 L_2)$-Lipschitz. Thus, the sequence of propagation steps remains 1-Lipschitz.

The final representation for an attribute, as per Eq.~\ref{eq7}, is a concatenation:
    \[
    \mathbf{h}_{x}^{*,k} = [\mathbf{h}_{x}^{k}; \tilde{\mathbf{h}}_{x}^{k}; \hat{\mathbf{h}}_{x}^{k}]
    \]
    Let $g(\mathbf{h}) = [\mathbf{h}; \mathcal{G}_{\text{intra}}(\mathbf{h}); \mathcal{G}_{\text{cross}}(\mathcal{G}_{\text{intra}}(\mathbf{h}))]$ be the operator that forms this vector. We can bound its Lipschitz constant:
$$
\begin{aligned}
\|g(\mathbf{h}) - g(\mathbf{h}')\|_2^2 &= \|\mathbf{h}-\mathbf{h}'\|_2^2 + \|\mathcal{G}_{\text{intra}}(\mathbf{h}) - \mathcal{G}_{\text{intra}}(\mathbf{h}')\|_2^2 \\ 
&\quad + \|\mathcal{G}_{\text{cross/intra}}(\mathbf{h}) - \mathcal{G}_{\text{cross/intra}}(\mathbf{h}')\|_2^2 \\
&\le \|\mathbf{h}-\mathbf{h}'\|_2^2 + 1^2 \|\mathbf{h}-\mathbf{h}'\|_2^2 + 1^2 \|\mathbf{h}-\mathbf{h}'\|_2^2 \\
&= 3 \|\mathbf{h}-\mathbf{h}'\|_2^2
\end{aligned}
$$
This implies that the concatenation operator is $\sqrt{3}$-Lipschitz. Subsequent operations like the gated fusion and the type-specific decoders are standard neural network layers that are also known to be Lipschitz continuous.
The entire mechanism is a composition of Lipschitz continuous functions and is therefore itself Lipschitz continuous. 
\end{proof}
This proves that small perturbations to the input features will result in a predictably bounded change in the final output, ensuring the model is robust.

\section{Comprehensive Experimental Configuration and Supplementary Results}
\subsection{Detailed Experimental Settings}
\label{app:detaledExpSetting}
\subsubsection{Evaluation Metrics}
\label{sec:appendix_evaluation_metrics}
Our evaluation metrics are tailored to the distinct characteristics of AIS data attributes. For continuous numerical values, we employ Mean Absolute Error (MAE) and Symmetric Mean Absolute Percentage Error (SMAPE):
\begin{equation}
    \textit{MAE} = \frac{1}{|\mathbf{\widetilde{M}}_{x_t}|}\sum_{x_t:\widetilde{m}_{x_t}=1}|x_t - \hat{x}_t|,
\end{equation}
where $x_t$ is the true value, $\hat{x}_t$ is the imputed value, and $|\mathbf{\widetilde{M}}_{x_t}|$ is the total number of imputed values in the target mask.
\begin{equation}
    \textit{SMAPE} = \frac{1}{|\mathbf{\widetilde{M}}_{x_t}|}\sum_{x_t:\widetilde{m}_{x_t}=1}\frac{|x_t - \hat{x}_t|}{(|x_t| + |\hat{x}_t|)/2},
\end{equation}
which effectively quantifies the relative deviation between imputed and true values. For categorical data, we utilize Accuracy (ACC):
\begin{equation}
    \textit{ACC} = \frac{1}{|\mathbf{\widetilde{M}}_{x_t}|}\sum_{x_t:\widetilde{m}_{x_t}=1}\mathbb{I}(x_t = \hat{x}_t), 
\end{equation}
where $\mathbb{I}(\cdot)$ is the indicator function that equals 1 when the prediction matches the ground truth and 0 otherwise.

For spatio-temporal attributes, we implement domain-specific evaluation approaches. 
For timestamp, we evaluate time intervals between consecutive records using MAE and SMAPE rather than absolute timestamps. 
For coordinate attributes (latitude and longitude), we compute the spherical distance between true and imputed coordinates in radians using haversine distance \cite{wiki_haversine_2023}.

\subsubsection{Baseline Methods}
\label{sec:appendix_baselines}
We briefly introduce the baselines as follows:
(1) \textbf{MEAN}: directly imputes missing attribute values in a sequence using the mean of each corresponding attribute. 
(2) \textbf{KNN}: imputes missing values by averaging the corresponding attributes from $k$ nearest records, with proximity determined by Euclidean distance in the feature space.
(3) \textbf{Lin-ITP}: performs linear interpolation of sequences for each attribute.
(4) \textbf{MF}: Matrix Factorization \cite{hastie2009elements} treats an AIS sequence as a matrix and employs iterative Singular Value Decomposition (SVD) to impute missing values.
(5) \textbf{TRMF}~\cite{yu2016temporal}: a temporal regularized matrix factorization method.
(6) \textbf{CSDI}~\cite{tashiro2021csdi}: a conditional score-based diffusion model for multi-variable time series imputation.
(7) \textbf{PriSTI}~\cite{liu_pristi_2023}: a conditional diffusion framework for spatiotemporal imputation.
(8) \textbf{ImputeFormer}~\cite{nie2024imputeformer}: a low rankness-induced Transformer model for spatiotemporal imputation.
(9) \textbf{Multi-task AIS}~\cite{nguyen2018multi}: a deep learning architecture that integrates recurrent neural networks with latent variable modeling, enabling simultaneous processing of trajectory imputation, anomaly detection, and vessel type identification tasks.
(10) \textbf{PG-DPM}~\cite{zhang2024long}: a physics-guided diffusion probabilistic model for long-term vessel trajectory imputation.

\subsubsection{Masking Strategies}
\label{sec:appendix_masking_strategies}
AIS data exhibits distinctive missing patterns across different time scales (as discussed in Section~\ref{sec:DataNotation}). To accurately simulate these real-world scenarios, we implement three targeted masking strategies with mask ratio $r$ that correspond to the characteristics of different attribute types and time scales:

\setlength{\parsep}{0pt}
\begin{itemize}[itemsep=2pt, leftmargin=10pt, parsep=-1pt]
\item \textbf{Point Masking}: Applied to the attributes in time scales 1 and 2 (i.e., $\mathbf{X}^1 = \{\lambda, \phi, \tau\}$ and $\mathbf{X}^2 = \{\psi, \theta, s\}$). For each attribute $x \in \mathbf{X}^1 \cup \mathbf{X}^2$ and time step $t$, we randomly mask individual values with probability $r$ according to:
\begin{equation}
\widetilde{m}_{x,t} = \begin{cases}
1 & \text{if } \text{Bernoulli}(r) = 1 \\
0 & \text{otherwise},
\end{cases}
\end{equation}
where $\widetilde{m}_{x,t}$ is the mask of attribute $x$ at time step $t$ (see Definition 7). This simulates sporadic absences due to signal interference or communication issues affecting high-frequency attributes, e.g., longitude ($\lambda$), latitude ($\phi$), timestamp ($\tau$), true heading angle ($\psi$), course over ground ($\theta$), and speed over ground ($s$).

\item \textbf{Block Masking}: Applied to the attributes in time scales 3 and 4 (i.e., $\mathbf{X}^3 = \{\eta\}$ and $\mathbf{X}^4 = \{\chi, d\}$). For each voyage segment $[t_{\text{start}}, t_{\text{end}}]$ and attribute $x \in \mathbf{X}^3 \cup \mathbf{X}^4$, we randomly mask continuous segments with probability $r$ according to:
\begin{equation}
\widetilde{m}_{x,t} = \begin{cases}
1 & \text{if } \text{Bernoulli}(r) = 1, \forall t \in [t_{\text{start}}, t_{\text{end}}] \\
0 & \text{otherwise},
\end{cases}
\end{equation}
where $t_{\text{start}}$ and $t_{\text{end}}$ define voyage segments based on operational phases identified by changes in draught ($d$) or hazardous cargo type ($\chi$). Within each segment, draught ($d$) and hazardous cargo type ($\chi$) are stable. This captures voyage-phase-related missing patterns for attributes like navigation status ($\eta$), hazardous cargo type ($\chi$), and draught ($d$).

\item \textbf{Entire Masking}: Applied to vessel-specific static attributes (i.e., $\mathbf{X}^5 = \{\ell, \beta, \kappa\}$). For each vessel and attribute $x \in \mathbf{X}^5$, we randomly mask the entire attribute sequence with probability $r$ according to:
\begin{equation}
\widetilde{m}_{x,t} = \begin{cases}
1 & \text{if } \text{Bernoulli}(r) = 1, \forall t \\
0 & \text{otherwise}
\end{cases}
\end{equation}
This replicates the systematic absence of vessel-specific data such as length ($\ell$), width ($\beta$), and vessel type ($\kappa$) frequently observed in AIS records.
\end{itemize}

\subsubsection{Noise Injection Strategies}
\label{sec:appendix_noise_injection_strategies}
A key consideration of noise injection strategies was to ensure that noise intensity parameter $\gamma$ functions as a \textbf{unified control} across all heterogeneous attribute types. To achieve this, we have designed our noise injection strategy to be data-driven and normalized against the intrinsic dynamics of each attribute. This ensures that a given level of $\gamma$ corresponds to a comparable degree of ``corruption'' for each attribute, relative to its typical behavior. Our type-specific noise injection strategies are as follows:

\noindent \textbf{Continuous Attributes.} 
To simulate random measurement errors, we add Gaussian noise. The magnitude of this noise is scaled by the attribute's own value, $x_n$, making the noise proportional to the attribute's magnitude. This method makes parameter $\gamma$ a fair, relative measure of corruption across attributes with different scales and value ranges. The noise is then added to the attribute value, $x_n$, as follows:
$$
x'_{n} = x_{n} + \mathcal{N}(0, (\gamma x_n)^2)
$$

\noindent \textbf{Spatio-temporal Attributes.} 
For coordinates, to realistically simulate positional drift from sources like GPS signal degradation, we link the noise magnitude to the vessel's own dynamics. By scaling the Gaussian noise with the standard deviation of the step-to-step coordinate change ($\sigma_{\Delta\lambda}, \sigma_{\Delta\phi}$), the noise becomes adaptive to the vessel's movement pattern. A fast-moving vessel will have larger positional uncertainty than a stationary one, which this method captures. The noise is then added to the coordinate value, $x_n$, as follows:
$$
\lambda' = \lambda + \mathcal{N}(0, (\gamma \sigma_{\Delta\lambda})^2)
$$
$$
\phi' = \phi + \mathcal{N}(0, (\gamma \sigma_{\Delta\phi})^2)
$$

For timestamp $\tau$, consistent with our model's method of predicting time intervals (as detailed in Section 4.4 and modeled as a temporal point process~\cite{lin2022exploring,omi2019fully}), we corrupt the time intervals ($\Delta\tau_i$) rather than the absolute timestamps. This directly tests the model's core mechanism for handling time. Noise is scaled by the standard deviation of these intervals ($\sigma_{\Delta\tau}$) to reflect typical reporting frequency fluctuations. A $\max(0, ...)$ function is applied as a necessary physical constraint to prevent time from moving backwards. The noise is then added to the time value $x_n$, as follows:
$$
\Delta\tau'_{i} = \max\left(0, \Delta\tau_i + \mathcal{N}(0, (\gamma \sigma_{\Delta\tau})^2)\right)
$$
$$
\tau'_{i} = \tau_{i-1} + \Delta\tau'_{i}
$$

\noindent \textbf{Cyclical Attributes.} To model the jitter commonly found in compass and directional sensor readings, we add Gaussian noise to the angular values. Scaling this noise by the standard deviation of the step-to-step angular change ($\sigma_{\Delta x_c}$) makes the noise magnitude adaptive to the vessel's turning dynamics. A modulo 360 operation is applied to ensure that the resulting values remain within the valid $[0, 360)$ range and correctly handle the wrap-around nature of cyclical data. The noise is then added to the angular value, $x_c$, as follows:
$$
x'_{c} = \left( x_{c} + \mathcal{N}(0, (\gamma \sigma_{\Delta x_c})^2) \right) \pmod{360}
$$

\noindent \textbf{Discrete Attributes.} 
Errors in categorical data are not numerical deviations but misclassifications (e.g., ``anchored'' is recorded as ``moored''). The most direct way to simulate this is via label flipping. Here, $\gamma$ defines the probability that a given label is flipped to another randomly chosen, valid label. While the role of $\gamma$ here (controlling corruption \textit{frequency}) differs from its role in other types (controlling corruption \textit{magnitude}), this preserves $\gamma$ as a unified parameter. It operates on the same $[0, 1]$ scale, where $\gamma=0$ means no noise and $\gamma=1$ implies maximum corruption for all attribute types. This method allows for a systematic and synchronized increase in domain-appropriate noise across the entire heterogeneous dataset. The noise is then added to the categorical value, $x_d$, as follows:
$$
x'_d = 
\begin{cases} 
    k \sim \text{Uniform}(C_{x_d} \setminus \{x_d\}) & \text{with probability } \gamma \\
    x_d & \text{with probability } 1-\gamma 
\end{cases}
$$

\begin{table*}[!htbp]
\small
\centering
\setlength{\tabcolsep}{0.8pt}
\caption{\warninge{Effectiveness analysis of attributes at different time scales.}}
\label{tab:ablation_attributes}
\begin{tabular}{c|l|c|cc|cc|cc|cc|c|c|cc|cc|cc|c}
\hline
\multicolumn{2}{c|}{\multirow{5}{*}{\makecell[c]{\textbf{Attribute}\\\textbf{Variant}}}} & \multicolumn{3}{c|}{\textbf{Scale 1}} & \multicolumn{6}{c|}{\textbf{Scale 2}} & \textbf{Scale 3} & \multicolumn{3}{c|}{\textbf{Scale 4}} & \multicolumn{5}{c}{\textbf{Scale 5}} \\
\cmidrule{3-20}
\multicolumn{2}{c|}{} & \makecell[c]{Coord. $(\lambda,\phi)$} & \multicolumn{2}{c|}{\makecell[c]{Time $\tau$}} & \multicolumn{2}{c|}{\makecell[c]{Head. $\psi$}} & \multicolumn{2}{c|}{\makecell[c]{Course $\theta$}} & \multicolumn{2}{c|}{\makecell[c]{Speed $s$}} & \makecell[c]{Nav. $\eta$} & \makecell[c]{Cargo $\chi$} & \multicolumn{2}{c|}{\makecell[c]{Draft $d$}} & \multicolumn{2}{c|}{\makecell[c]{Length $\ell$}} & \multicolumn{2}{c|}{\makecell[c]{Width $\beta$}} & \makecell[c]{Type $\kappa$} \\
\cmidrule{3-20}
\multicolumn{2}{c|}{} & Dist. & MAE & SMAPE & MAE & SMAPE & MAE & SMAPE & MAE & SMAPE & ACC & ACC & MAE & SMAPE & MAE & SMAPE & MAE & SMAPE & ACC \\
\midrule
\multirow{6}{*}{\rotatebox{90}{AIS-DK}} & w/o Scale 1 & - & - & - & \warninge{0.070} & \warninge{0.044} & \warninge{0.168} & \warninge{0.083} & \warninge{0.238} & \warninge{0.273} & \warninge{0.814} & \warninge{0.729} & \warninge{1.316} & \warninge{0.170} & \warninge{42.236} & \warninge{0.216} & \warninge{6.244} & \warninge{0.203} & \warninge{0.592} \\
& w/o Scale 2 & \warninge{1.315e-4} & \warninge{25.762} & \warninge{0.157} & - & - & - & - & - & - & \warninge{0.886} & \warninge{0.772} & \warninge{1.132} & \warninge{0.141} & \warninge{35.096} & \warninge{0.184} & \warninge{4.982} & \warninge{0.182} & \warninge{0.653} \\
& w/o Scale 3 & \warninge{1.223e-4} & \warninge{22.631} & \warninge{0.137} & \warninge{0.054} & \warninge{0.039} & \warninge{0.145} & \warninge{0.075} & \warninge{0.207} & \warninge{0.256} & - & \warninge{0.772} & \warninge{1.140} & \warninge{0.141} & \warninge{35.036} & \warninge{0.183} & \warninge{4.962} & \warninge{0.180} & \warninge{0.659} \\
& w/o Scale 4 & \warninge{1.182e-4} & \warninge{22.131} & \warninge{0.136} & \warninge{0.052} & \warninge{0.038} & \warninge{0.144} & \warninge{0.074} & \warninge{0.206} & \warninge{0.262} & \warninge{0.886} & - & - & - & \warninge{35.442} & \warninge{0.189} & \warninge{5.062} & \warninge{0.183} & \warninge{0.647} \\
& w/o Scale 5 & \warninge{1.149e-4} & \warninge{22.254} & \warninge{0.136} & \warninge{0.056} & \warninge{0.043} & \warninge{0.146} & \warninge{0.078} & \warninge{0.208} & \warninge{0.263} & \warninge{0.888} & \warninge{0.769} & \warninge{1.138} & \warninge{0.142} & - & - & - & - & - \\
& \textsf{MH-GIN} & \warninge{\textbf{1.112e-4}} & \warninge{\textbf{21.413}} & \warninge{\textbf{0.127}} & \warninge{\textbf{0.049}} & \warninge{\textbf{0.036}} & \warninge{\textbf{0.141}} & \warninge{\textbf{0.071}} & \warninge{\textbf{0.198}} & \warninge{\textbf{0.252}} & \warninge{\textbf{0.904}} & \warninge{\textbf{0.798}} & \warninge{\textbf{1.017}} & \warninge{\textbf{0.126}} & \warninge{\textbf{34.192}} & \warninge{\textbf{0.178}} & \warninge{\textbf{4.896}} & \warninge{\textbf{0.175}} & \warninge{\textbf{0.669}} \\
\midrule
\multirow{6}{*}{\rotatebox{90}{AIS-US}} & w/o Scale 1 & - & - & - & \warninge{0.131} & \warninge{0.124} & \warninge{0.491} & \warninge{0.237} & \warninge{0.467} & \warninge{1.323} & \warninge{0.601} & \warninge{0.529} & \warninge{2.244} & \warninge{0.346} & \warninge{41.441} & \warninge{0.323} & \warninge{5.858} & \warninge{0.370} & \warninge{0.402} \\
& w/o Scale 2 & \warninge{9.218e-5} & \warninge{17.493} & \warninge{0.035} & - & - & - & - & - & - & \warninge{0.637} & \warninge{0.549} & \warninge{1.926} & \warninge{0.320} & \warninge{38.929} & \warninge{0.279} & \warninge{5.403} & \warninge{0.214} & \warninge{0.432} \\
& w/o Scale 3 & \warninge{8.712e-5} & \warninge{15.841} & \warninge{0.018} & \warninge{0.113} & \warninge{0.079} & \warninge{0.469} & \warninge{0.187} & \warninge{0.374} & \warninge{1.189} & - & \warninge{0.552} & \warninge{1.939} & \warninge{0.360} & \warninge{38.953} & \warninge{0.279} & \warninge{5.405} & \warninge{0.215} & \warninge{0.429} \\
& w/o Scale 4 & \warninge{8.729e-5} & \warninge{15.875} & \warninge{0.020} & \warninge{0.122} & \warninge{0.091} & \warninge{0.472} & \warninge{0.187} & \warninge{0.378} & \warninge{1.191} & \warninge{0.639} & - & - & - & \warninge{39.114} & \warninge{0.282} & \warninge{5.414} & \warninge{0.216} & \warninge{0.427} \\
& w/o Scale 5 & \warninge{8.619e-5} & \warninge{15.854} & \warninge{0.017} & \warninge{0.116} & \warninge{0.083} & \warninge{0.470} & \warninge{0.188} & \warninge{0.376} & \warninge{1.188} & \warninge{0.637} & \warninge{0.551} & \warninge{1.953} & \warninge{0.341} & - & - & - & - & - \\
& \textsf{MH-GIN} & \warninge{\textbf{8.281e-5}} & \warninge{\textbf{15.146}} & \warninge{\textbf{0.016}} & \warninge{\textbf{0.104}} & \warninge{\textbf{0.072}} & \warninge{\textbf{0.461}} & \warninge{\textbf{0.184}} & \warninge{\textbf{0.368}} & \warninge{\textbf{1.182}} & \warninge{\textbf{0.651}} & \warninge{\textbf{0.559}} & \warninge{\textbf{1.899}} & \warninge{\textbf{0.314}} & \warninge{\textbf{38.263}} & \warninge{\textbf{0.273}} & \warninge{\textbf{5.382}} & \warninge{\textbf{0.213}} & \warninge{\textbf{0.446}} \\
\hline
\end{tabular}
\end{table*}
\subsection{Attributes Ablation}
We examined the importance of different attribute types and their interactions in Table \ref{tab:ablation_attributes}, providing insights into how each attribute category contributes to the overall imputation performance. 

We have following observations:
1) Attributes at time scale 1 significantly impact the imputation accuracy of other attributes. When real-time attributes are added, MAE and SMAPE reduce substantially in both datasets, and the error rate of discrete attributes decreases significantly, demonstrating that positional and temporal information provides essential context for movement-related attributes.
2) Attributes at time scale 2 play a crucial role in accurately imputing vessel positions. Their addition reduces coordinate distance errors and time interval MAE and SMAPE substantially, suggesting that directional and speed information substantially constrains possible vessel positions.
3) Navigation status attribute, despite their lower update frequency, contributes meaningfully to the model's performance. Their addition reduces cargo status imputation accuracy error and draft errors, highlighting their value in characterizing vessel operational patterns.
4) Attributes at time scale 4 provide important context for static vessel characteristics. With them, vessel type imputation accuracy error decreases, and dimension imputation errors decrease as well, indicating their role in capturing vessel identity features. 
5) Attributes at time scale 5 offer valuable anchoring information for dynamic imputations. Their addition results in coordinate errors decreasing and heading imputation MAE reducing substantially, showing that fixed vessel characteristics constrain the space of plausible movement patterns.

Overall, the consistent performance improvement observed when adding any attribute category confirms the effectiveness of our multi-scale attribute modeling approach. The bidirectional influence between attributes of different update frequencies demonstrates that \textsf{MH-GIN} successfully leverages cross-scale dependencies, with each attribute type providing complementary information that enhances the entire imputation process.

\end{document}